\documentclass{article} %

\usepackage{colm2024_conference}

\usepackage{footmisc}
\usepackage{booktabs}
\usepackage{graphicx}
\usepackage{enumitem}
\usepackage{wrapfig}
\usepackage{algorithm}
\usepackage{algpseudocode}
\usepackage[T1]{fontenc}

\usepackage{lmodern}
\usepackage{pifont}
\usepackage{amsmath}
\usepackage{bm}

\usepackage{microtype}
\usepackage{amsmath}
\usepackage{colortbl}
\usepackage[utf8]{inputenc}
\definecolor{lightgray}{rgb}{0.9,0.9,0.9}
\usepackage{caption}
\usepackage{subcaption}
\usepackage{xcolor}
\usepackage{setspace}
\usepackage{url}
\usepackage{multirow}
\usepackage{colortbl}

\usepackage{tabularx}
\usepackage{blindtext}
\usepackage{pgfplots}
\pgfplotsset{compat=1.18} 
\usepackage{tikz}
\usetikzlibrary{er,positioning,bayesnet}
\usepackage{makecell}
\usepackage{tipa}
\usepackage{siunitx}
\usepackage{nicefrac}
\usepackage{tocloft}
\usepackage{listings}
\usepackage[raster,skins]{tcolorbox} %
\usepackage{xltabular}
\usepackage{adjustbox}
\usepackage{xurl}
\usepackage{marvosym}
\usepackage{algpseudocode}

\usepackage{amsmath}
\usepackage{amssymb}
\usepackage{mathtools}
\usepackage{amsthm}
\usepackage{enumitem}

\usepackage[table]{xcolor}

\definecolor{outlier1}{RGB}{250,252,255}
\definecolor{outlier2}{RGB}{238,244,250}
\definecolor{outlier3}{RGB}{226,236,245}
\definecolor{outlier4}{RGB}{210,225,240}
\definecolor{outlier5}{RGB}{190,210,230}
\definecolor{outlier6}{RGB}{170,195,220}
\definecolor{outlier7}{RGB}{150,180,210}

\definecolor{loss1}{RGB}{250,253,250}
\definecolor{loss2}{RGB}{242,250,240}
\definecolor{loss3}{RGB}{230,245,230}
\definecolor{loss4}{RGB}{215,235,210}
\definecolor{loss5}{RGB}{200,225,200}
\definecolor{loss6}{RGB}{185,215,185}
\definecolor{loss7}{RGB}{170,205,170}

\definecolor{gap_red1}{RGB}{255,245,205}
\definecolor{gap_red2}{RGB}{255,230,190}
\definecolor{gap_red3}{RGB}{255,210,170}
\definecolor{gap_green1}{RGB}{235,250,205}
\definecolor{gap_green2}{RGB}{220,245,190}
\definecolor{gap_green3}{RGB}{200,235,170}

\author{\textbf{Qwen Team}
}
\title{
On the Design of Qwen3.8-Next Architecture:
Evaluation, Efficiency, and Training Stability
}
\begin{document}

\maketitle

\begin{abstract}

We describe the architecture and ablations of \textbf{Qwen3.8-Flash-Next}, a sparse mixture-of-experts model with 125B parameters, 6B activated per token, and additional 51B parameters of n-gram embedding tables held off the accelerator.
On fourteen pre-training benchmarks the model leads the 397B-A17B predecessor on eight and trails it on the rest by at most 2.6 points, at \textbf{1/3} the activated parameters, \textbf{1/3} the training tokens, and roughly \textbf{1/9} the training FLOPs.
Token mixing uses a layer-wise hybrid of \textbf{Gated DeltaNet (GDN)} and global attention, with one full-attention layer in every four; at continued-pretraining time those full-attention layers are replaced by \textbf{Qwen Sparse Attention (QSA)}, which scores context at micro-block granularity with a compressed lightweight indexer. The residual stream is widened to four branches and read through an elementwise gate, a design we call the \textbf{Gated Residual (GR)}. 
Capacity is added outside the backbone by a single \textbf{n-gram embedding layer} whose tables are prefetched from host memory.
We evaluate every candidate change along three axes: loss together with downstream benchmarks; the cost of the change in training, prefill and decode; and its effect on the optimal hyperparameters and training stability. 
Loss and downstream accuracy do not always move together: enlarging the n-gram vocabulary lowers loss monotonically while downstream accuracy saturates.
The architecture and the \textbf{Muon} optimizer together shift the optimal learning rate and batch size upwards, render batch-size warmup unnecessary, and substantially improve stability under stress tests.
Loss, benchmarks, efficiency and stability form one design problem. 
Solved jointly, they yield a recipe that is simultaneously \textit{more efficient, more capable and more stable}.
\end{abstract}

\section{Introduction}


Qwen3.8-Flash-Next is a sparse mixture-of-experts model with 125B total parameters, 6B activated per token, and additional 51B parameters of n-gram embedding tables held off the accelerator.
The design goal is to retain the quality of the
previous generation's 397B-A17B flagship~\citep{qwenteam2026qwen35} at a fraction of its compute budget.
On fourteen pre-training benchmarks spanning knowledge, STEM, reasoning, coding and multilingual ability, the resulting base model leads that predecessor on eight and trails it on the remaining six by at most $2.6$ points (Tab.~\ref{tab:qwen37-comparison}), while activating roughly a third as many parameters per token and training on roughly a third as many tokens, for about a ninth of the training FLOPs.

An architectural change touches three things at once: what the model can do on downstream tasks, what it costs to train and to serve, and whether the training run remains optimal and stable at scale. 
We therefore evaluate every candidate
change along three axes: loss together with downstream benchmarks; 
the cost of the change in training, prefill and decode; and its effect on the optimal hyperparameters and on training stability. 
Loss, benchmarks, efficiency and stability form one design problem, and we report each on its own terms.
Throughout this report, we highlight where the three axes disagree and the design choices those disagreements led to.

Four architectural components carry the design, each addressing a distinct bottleneck. 
Token mixing uses a layer-wise hybrid of Gated
DeltaNet (GDN)~\citep{yang2024gated} and global attention: the recurrent layers compress the prefix into a fixed-size state at linear cost, while one full-attention layer in every four retains the direct token-level retrieval that no finite-state memory reproduces exactly (\S\ref{sec:gdn-hybrid}). 
At continued-pretraining time those full-attention layers are replaced by Qwen Sparse Attention (QSA), which follows the sparse-attention route of~\citet{liu2025deepseek} but scores context at micro-block granularity with a compressed lightweight indexer, so that the cost of indexing itself falls with sequence length. 
The residual stream is widened to four branches~\citep{baykal2023altup,zhu2024hyper} and read through an elementwise gate, a design we call Gated Residual (GR) (\S\ref{sec:residual}): widening adds capacity to the residual path, and the gate decides how that capacity is spent, while also supplying the rescaling that keeps training stable. 
Capacity is further added outside the backbone by a single n-gram embedding layer~\citep{google2025gemma3n,cheng2026engram} whose tables are prefetched from host memory (\S\ref{sec:ngram-emb}), scaling parameter count with negligible additional per-token FLOPs and latency.

\paragraph{Evaluation.}
Loss and downstream accuracy do not always move together, and we observe
disagreements in both directions. Enlarging the n-gram vocabulary lowers loss
monotonically while downstream accuracy saturates, and under a fixed parameter
budget the loss optimum diverges from the accuracy optimum
(Tab.~\ref{tab:ngram-vocab-size}, Tab.~\ref{tab:ngram-vocab-size-samebudget}).
Conversely, predicting the residual read and write weights from the residual
state yields only a marginal loss reduction but a clear benchmark gain
(\S\ref{sec:residual-ablation}). 
Other disagreements surface late: restricting each block to the two highest-gated residual branches is almost free in pre-training loss yet degrades with further training (\S\ref{sec:residual-efficiency}), and removing positional encoding from the full-attention layers is indistinguishable during pre-training but affects generation quality at later stages (\S\ref{sec:gdn-hybrid}). 
In the sections that follow, where loss and benchmarks move together we report loss alone, for brevity.

\begin{figure}[t]
  \centering
  \vskip -0.2in
  \includegraphics[pagebox=cropbox,width=0.8\linewidth]{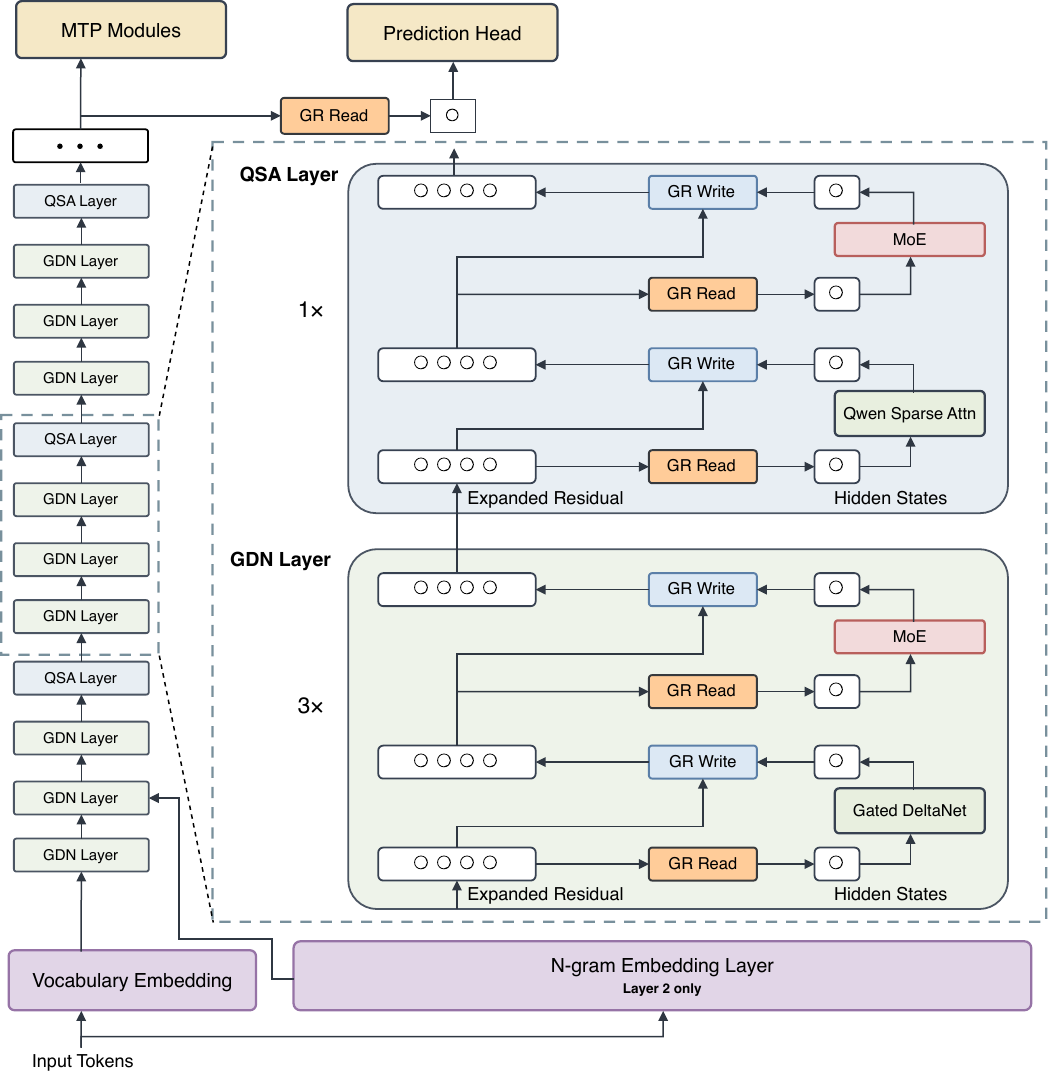}
  \vskip -0.1in
\caption{\textbf{Qwen3.8-Flash-Next architecture.} Token mixing alternates three GDN layers with one QSA layer per block of four. Every sublayer reads and writes through GR, which widens the residual stream and gates the read elementwise. 
An n-gram embedding layer at Layer 2 scales capacity off the accelerator via host-memory prefetching. 
The MTP module reuses QSA indices across speculative decoding steps.
}
  \label{fig:whole_arch}
\end{figure}

\paragraph{Efficiency.}
We evaluate cost separately in training, prefill and decode. \textbf{In training}, FlashQLA achieves a $2$--$3\times$ forward and roughly $2\times$ backward speedup over the Triton baseline on GPUs (\S\ref{sec:gdn-efficiency}). 
Muon introduces its own engineering costs: its per-parameter FLOPs depend on matrix shape rather than parameter count, so
the data-parallel gradient buffer is repartitioned by estimated orthogonalization cost; and its step fragments into many small kernels once fused parameters are split, so the step is captured in a CUDA graph (\S\ref{sec:optimizer-impl}).
We set the Newton--Schulz iteration to 8 steps, favouring the additional stability under stress. 
At inference, \textbf{prefill} is dominated by attention over the whole context, which QSA addresses by compressing the key sequence by a factor $r$, reducing indexer cost from $O(n^2)$ to $O(n^2/r)$. 
\textbf{Decode} is dominated by memory traffic, which is why the GDN layers keep a fixed-size recurrent state, the GR drops the branch-mixing operator $H_{\mathrm{res}}$, and the
residual state supports FP8 storage. 
At a context length of 1M, QSA is $7.6\times$ faster than dense attention in prefill and $4.9\times$ faster in decode at the kernel level.

\paragraph{Optimization.}
Muon~\citep{jordan2024muon} is applied to the two-dimensional weights that act as linear maps; the input and n-gram embeddings, output head, MoE router and the
low-rank projections of GR stay on AdamW, where orthogonalization is either
impractical or unhelpful (\S\ref{sec:optimizer}). 
Fused parameters are split before orthogonalization, since orthogonalizing a concatenated matrix mixes
singular directions across unrelated sub-blocks
(\S\ref{sec:optimizer-split}). 
The new architecture and optimizer also shift
the optimal hyperparameters, so we refit the scaling
law~\citep{kaplan2020scaling} used for the Qwen3.5 series
(\S\ref{sec:hpscaling}).
The new scaling law predicts a larger batch size and learning rate; both predictions are verified separately and confirmed. 
The larger batch size improves parallel throughput at scale, and the larger learning rate improves convergence. 
Ramping the batch size over early training ends no better than starting at the target and costs $18.8\%$ more optimizer steps, so we do not use it (\S\ref{sec:hpscaling-warmup}).

\paragraph{Training Stability.}
We verify training stability under \textbf{stress tests} that reproduce large-scale instabilities at moderate model scale by raising the learning rate~\citep{wortsman2023smallscale} and keeping a constant learning rate. 
\textit{The criterion is that the new recipe must be at least as stable as the generation it replaces under equal stress.}
At four times the optimal learning rate, the previous structure spikes frequently, whereas the new recipe remains stable throughout (\S\ref{sec:stability}). 
Isolating the gate in GR on a single-variable pair confirms it as a key contributor to the stability margin over the Qwen3.5 architecture. 
As a direct result of these stability enhancements, the full-scale training of Qwen3.8-Flash-Next proceeded smoothly without a single loss spike or anomalous fluctuation in gradient norms, without relying on explicit clipping methods such as qk-clip~\citep{kimi2025k2} or SwiGLU-clip~\citep{agarwal2025gpt}.

\paragraph{Organization.}
\S\ref{sec:gdn-hybrid} and \S\ref{sec:qsa} describe token mixing.
\S\ref{sec:residual} covers the gated residual, \S\ref{sec:ngram-emb} the
n-gram embedding layer, \S\ref{sec:optimizer} the optimizer,
\S\ref{sec:hpscaling} the hyperparameter scaling and \S\ref{sec:stability}
the stability stress test. Evaluation of the resulting base and post-trained
models follows.



\section{Model Architecture}

\subsection{Attention}

\subsubsection{GDN Hybrid Architecture}
\label{sec:gdn-hybrid}

\begin{figure}[H]
\centering
\includegraphics[pagebox=cropbox,width=0.52\linewidth]{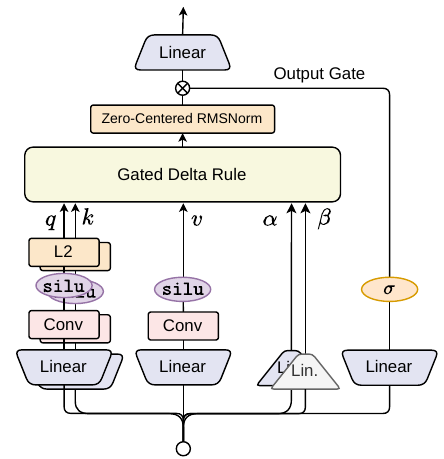}
\caption{The Gated DeltaNet token mixer. The projected query, key, and value streams pass through short causal convolutions; queries and keys are L2-normalized before the gated delta recurrence. The decay gate $\alpha_t$ and write gate $\beta_t$ control the recurrent update, while a sigmoid output gate modulates the zero-centered RMS-normalized output.}
\label{fig:gdn-module}
\end{figure}

\paragraph{Motivation.}
\label{sec:gdn-motivation}

Full self-attention provides direct content-based access to every preceding token, but its token-mixing cost grows quadratically with sequence length and its key--value (KV) cache grows linearly during autoregressive generation~\citep{vaswani2017attention}. Sliding-window attention (SWA) replaces global access with a bounded local receptive field, reducing both computation and cache consumption. However, information outside the window can only propagate indirectly through depth. This creates a tension between efficient local processing and persistent content-dependent memory.

We address this tension with a layer-wise hybrid of Gated DeltaNet (GDN)~\citep{yang2024gated} and global attention. GDN compresses the prefix into a fixed-size recurrent state and updates that state according to the current content, while the interleaved global-attention layers retain direct token-level retrieval that is difficult for any finite-state recurrent memory to reproduce exactly.

This design choice is consistent with the architecture ablation reported in Tab.~\ref{tab:gdn-main-results}. Relative to the full-attention Transformer baseline, the GDN-hybrid model improves 8 of the 9 selected benchmarks. Relative to the SWA-hybrid, it is stronger on seven of the nine benchmarks. These results motivate a hybrid design, but they do not by themselves isolate which architectural component causes each improvement.

\paragraph{Gated Delta Recurrence.}
\label{sec:gdn-recurrence}

Linear attention can be interpreted as a fast-weight memory that stores key--value associations in a matrix state~\citep{schlag2021linear}. For each head, let $\bm{q}_t,\bm{k}_t\in\mathbb{R}^{d_k}$ and $\bm{v}_t\in\mathbb{R}^{d_v}$. Following the implementation convention, GDN maintains a state $\bm{S}_t\in\mathbb{R}^{d_k\times d_v}$, which is the transpose of the state convention used in the original formulation, and applies the gated delta rule~\citep{yang2024gated}:
\begin{align}
\widetilde{\bm{S}}_{t-1} &= \alpha_t \bm{S}_{t-1}, \\
\bm{e}_t &= \bm{v}_t - \widetilde{\bm{S}}_{t-1}^{\top}\bm{k}_t, \\
\bm{S}_t &= \widetilde{\bm{S}}_{t-1}
    + \beta_t\bm{k}_t\bm{e}_t^{\top}, \\
\bm{y}_t &= \bm{S}_t^{\top}\bm{q}_t,
\end{align}
where $\alpha_t\in(0,1)$ is a data-dependent decay and $\beta_t\in(0,1)$ controls the delta update. Equivalently,
\begin{equation}
\bm{S}_t
= \alpha_t\left(\bm{I}-\beta_t\bm{k}_t\bm{k}_t^{\top}\right)\bm{S}_{t-1}
+ \beta_t\bm{k}_t\bm{v}_t^{\top}.
\end{equation}

The two gates play complementary roles. The decay $\alpha_t$ globally controls the lifetime of the existing state, whereas the delta term first estimates the value already associated with $\bm{k}_t$ and writes only the residual error. Consequently, repeated or similar keys update an existing association instead of accumulating unbounded outer products. This targeted erase-and-write operation distinguishes GDN from purely additive linear attention.

\paragraph{GDN Parameterization.}
\label{sec:gdn-parameterization}

Given the normalized residual-stream input $\bm{x}_t\in\mathbb{R}^{d}$, GDN computes content features using learned projections followed by a short depthwise causal convolution:
\begin{align}
\bm{q}_t &= \operatorname{L2Norm}\left(\operatorname{SiLU}\left(\operatorname{ShortConv}(\bm{W}_q\bm{x}_t)\right)\right), \\
\bm{k}_t &= \operatorname{L2Norm}\left(\operatorname{SiLU}\left(\operatorname{ShortConv}(\bm{W}_k\bm{x}_t)\right)\right), \\
\bm{v}_t &= \operatorname{SiLU}\left(\operatorname{ShortConv}(\bm{W}_v\bm{x}_t)\right).
\end{align}
Short convolution supplies an explicit local inductive bias before information is compressed into the recurrent state. L2 normalization bounds the magnitudes of $\bm{q}$/$\bm{k}$ and stabilizes the rank-one delta transition.

The write strength and decay are parameterized as
\begin{align}
\beta_t &= \sigma(\bm{W}_{\beta}\bm{x}_t), \\
\alpha_t &= \exp\left[-\exp(\bm{A})
    \operatorname{softplus}(\bm{W}_{\alpha}\bm{x}_t+\bm{b}_{\alpha})\right].
\end{align}
After applying the recurrence independently across heads, the head outputs are normalized and modulated by an input-dependent output gate:
\begin{equation}
\label{eq:gdn-output-gate}
\bm{o}_t = \bm{W}_o\left[
\sigma(\bm{W}_z\bm{x}_t)
\odot \operatorname{RMSNorm}(\bm{y}_t)
\right].
\end{equation}
Unlike the original GDN, which uses a SiLU output gate, we use the bounded sigmoid gate in Equation~\eqref{eq:gdn-output-gate} and observe consistent improvements across our experiments. 
Following Qwen3-Next, we adopt zero-centered RMSNorm to constrain the growth of RMSNorm weights. The same formulation is consistently applied to all other RMSNorm layers used throughout the model.
Figure~\ref{fig:gdn-module} summarizes the complete GDN token-mixing path.

At the model level, our hybrid architecture retains rotary position embeddings (RoPE)~\citep{rope} in its full-attention layers. 
RoPE and a NoPE variant without positional encoding show little difference during pretraining, but \textit{the NoPE variant exhibits a substantially higher rate of endless generation after post-training and is therefore more likely to fail to terminate.}
We place one such full-attention layer in every four layers, with the other three layers using GDN. This schedule strikes a favorable balance between efficiency and quality, while periodic full attention is particularly important for long-context performance.

\paragraph{Kernel Efficiency.}
\label{sec:gdn-efficiency}

For efficiency, we optimize the GDN kernel with FlashQLA, a TileLang-based fused linear-attention kernel library. Across multiple settings on NVIDIA GPUs, FlashQLA achieves a $2$--$3\times$ forward speedup and an approximately $2\times$ backward speedup over the FLA Triton kernel~\citep{yang2024fla}. The implementation and benchmarks are available at \url{https://github.com/QwenLM/FlashQLA}.

\paragraph{Architecture Ablation.}
\label{sec:gdn-ablation}

We compare three checkpoints evaluated by the same evaluation pipeline: a full-attention Transformer, an SWA hybrid, and the GDN hybrid. Both hybrid variants use one full-attention layer in every four layers; the remaining token-mixing layers use SWA or GDN, respectively, with a window size of 128 for the SWA layers. Each checkpoint corresponds to a 28-layer 25B-A3B MoE model based on the Qwen3.5 architecture, pretrained first on 400B tokens with a 4K context length, and subsequently on 80B tokens with a 32K context length. We report knowledge results on MMLU~\citep{mmlu}, MMLU-Pro~\citep{wang2024mmlu}, and SuperGPQA~\citep{supergpqa}; STEM results on MATH~\citep{hendrycks2021measuring} and GSM8K~\citep{cobbe2021training}; reasoning results on BBH~\citep{suzgun2022bbh}; multilingual results on MMMLU~\citep{openai2024mmmlu}; and code results on EvalPlus and on MultiPL-E~\citep{cassano2023multipl}. The results are reported in Table~\ref{tab:gdn-main-results}.

\begin{table*}[t]
\centering
\small
\setlength{\tabcolsep}{3.5pt}
\caption{Architecture comparison. All values are percentages and higher is better. EvalPlus and MultiPL-E are reported with pass@1-style aggregate fields. Avg. is the unweighted arithmetic mean across the nine benchmarks. Bold denotes the best result in each column.}
\label{tab:gdn-main-results}
\resizebox{\textwidth}{!}{%
\begin{tabular}{lrrrrrrrrrr}
\toprule
& \multicolumn{3}{c}{Knowledge} & \multicolumn{2}{c}{STEM} & Reasoning & Multilingual & \multicolumn{2}{c}{Code} & \\
\cmidrule(lr){2-4} \cmidrule(lr){5-6} \cmidrule(lr){7-7} \cmidrule(lr){8-8} \cmidrule(lr){9-10}
Architecture & MMLU & MMLU-Pro & SuperGPQA & MATH & GSM8K & BBH & MMMLU & EvalPlus & MultiPL-E & Avg. \\
\midrule
Full attention
& 62.65 & 37.59 & 21.76 & 49.40 & 75.13 & 63.78 & 47.74 & 51.01 & 39.73 & 49.87 \\
SWA hybrid
& \textbf{66.30} & 40.67 & 22.45 & 45.48 & 74.22 & 65.88 & 51.33 & \textbf{52.12} & 41.93 & 51.15 \\
GDN hybrid
& 66.26 & \textbf{42.82} & \textbf{23.45} & \textbf{53.98} & \textbf{77.07} & \textbf{68.72} & \textbf{54.83} & 49.71 & \textbf{47.48} & \textbf{53.81} \\
\bottomrule
\end{tabular}
}
\end{table*}

The GDN hybrid improves over the Transformer on eight of the nine selected benchmarks and exceeds the SWA hybrid on seven. It achieves the best result on seven benchmarks and the highest overall average, while the SWA hybrid is marginally higher on MMLU and stronger on EvalPlus.

\subsubsection{Qwen Sparse Attention}
\label{sec:qsa}

\paragraph{Motivation.}
By selectively attending to important context, sparse attention alleviates the quadratic computational bottleneck of softmax attention in long-context scenarios.
Recently, DSA~\citep{liu2025deepseek} has achieved considerable inference speedups using a lightweight indexer to generate token-level sparse masks.
However, as sequence length increases, the overhead of its $O(n^2)$ indexer remains non-negligible.

\begin{figure}[h]
    \centering
    \includegraphics[width=\linewidth]{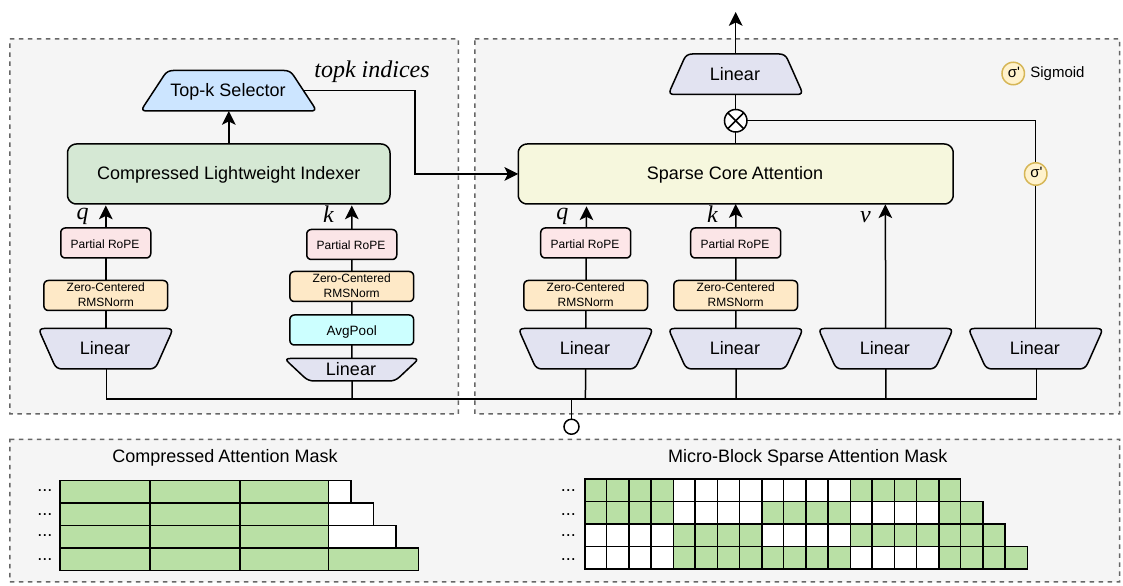}
    \caption{Overview of Qwen Sparse Attention (QSA). The QSA indexer (left) uses a compressed causal attention mask to score key blocks and select the top-$k$ indices. 
    These indices are expanded into a micro-block sparse attention mask for sparse core attention (right).}
    \label{fig:qsa_arch}
\end{figure}

To reduce this indexing cost, Qwen3.8-Flash-Next adopts Qwen Sparse Attention (QSA).
Specifically, QSA employs a lightweight indexer that compresses the sequence into micro-block representations, estimates their importance, and selects the most relevant context for attention computation.
This design reduces indexing overhead on long inputs while balancing task performance and inference efficiency.
Compared with methods that share indices across layers, the within-layer sequence compression in QSA relies less on cross-layer similarity, making it naturally suited to hybrid architectures.

Figure \ref{fig:qsa_arch} illustrates the QSA architecture. A compressed lightweight indexer first estimates context importance at micro-block granularity and then guides sparse computation in the core attention module.

\paragraph{Compressed Lightweight Indexer.}
Given an input hidden state $\mathbf{x}_i$, the indexer adopts an MQA~\citep{shazeer2019fast} with $H$ query heads and one shared key head.
It first applies independent lightweight projections:
\begin{equation}
    \widehat{\mathbf{q}}^h_i
    = \operatorname{RMSNorm}(\mathbf{W}^h_Q\mathbf{x}_i),
    \qquad
    \mathbf{k}_i=\mathbf{W}_K\mathbf{x}_i.
\end{equation}
To reduce the sequence length, keys are partitioned into non-overlapping blocks of $r$ tokens and compressed by average pooling.
Denoting the starting position of block $b$ by $p_b=b\cdot r$, the corresponding compressed key is
\begin{equation}
    \widehat{\mathbf{k}}_b
    = \operatorname{RMSNorm}\left(
        \operatorname{AvgPool}
        (\mathbf{k}_{p_b:p_b+r-1})
      \right),
    \qquad
    0\le b<\left\lfloor\frac{n}{r}\right\rfloor.
    \label{eq:qsa-avg-pooling}
\end{equation}
For positional encoding, the indexer applies partial RoPE.
Specifically, partial RoPE is applied to 64 of the 128 dimensions in each indexer head, matching the rotary dimension used in the core attention module. 
As shown in Eq.~\eqref{eq:qsa-avg-pooling}, key compression is performed before positional encoding.
Each block is therefore first summarized into a content representation and then assigned a single block-level position.
This ordering avoids averaging token representations with different rotary phases.
Each query retains its token position $i$, whereas each compressed key is assigned the starting position $p_b$ of its block:
\begin{equation}
    \mathbf{q}_i^h
    = \operatorname{PRoPE}(\widehat{\mathbf{q}}_i^h, i),
    \qquad
    \bar{\mathbf{k}}_b
    = \operatorname{PRoPE}(\widehat{\mathbf{k}}_b, p_b).
    \label{eq:qsa-partial-rope}
\end{equation}
Block-level importance scores $\mathbf{I}$ are then obtained through
block-causal scoring.
For query token $i$ and compressed block $b$,
ReLU-activated query--key similarities are summed over all indexer heads:
\begin{equation}
    I_{ib}
    =
    \left\{
    \begin{array}{cl}
        \displaystyle
        \sum_{h=1}^{H}
        \operatorname{ReLU}\left(
        \left\langle\mathbf{q}_i^h,\bar{\mathbf{k}}_b\right\rangle
        \right),
        & p_b+r-1\le i,\\[6pt]
        -\infty,
        & \text{otherwise},
    \end{array}
    \right.
    \label{eq:qsa-index-score}
\end{equation}
This block-causal condition allows each query to score only blocks that have been fully observed.
Given a token budget $K$, each query selects the highest-scoring compressed blocks.
Since each block contains $r$ tokens, the block budget is
$K_B=\lceil K/r\rceil$:
\begin{equation}
    \mathcal{B}_i
    = \operatorname{TopK}_{K_B}\left(\{I_{ib}\}_b\right),
    \qquad
    K_B=\left\lceil\frac{K}{r}\right\rceil.
    \label{eq:qsa-topk-blocks}
\end{equation}
Selected blocks are then expanded to their original token indices and
truncated to the budget $K$.
Together with the tokens in the final incomplete
block, which are always included, they form the final set used for core attention computation.

\paragraph{Training Details.}
QSA is introduced during the continued pretraining (CPT) stage of
Qwen3.8-Flash-Next with a sequence length of 256K tokens.
The training procedure consists of two stages: dense distillation and sparse training.
\begin{itemize}[leftmargin=1.5em]
\item \textbf{Stage 1: Dense Distillation.}
We first distill the full-sequence attention distribution of the backbone into the indexer.
The token-level teacher distribution is obtained by summing the softmax attention distributions over all teacher heads and applying $L_1$ normalization.
Denoting the resulting probability from query $i$ to token $j$ by $a_{ij}$, the full token-level distribution is $\mathbf{a}_i\in\mathbb{R}^{n}$.
Following prior work~\citep{gao2024seerattention,wang2026proxyattn},
we apply max pooling to align this distribution with the block-level indexer scores, 
thereby preserving salient token-level signals that could otherwise be diluted during aggregation:
\begin{equation}
\bar{a}_{ib}
= \operatorname{MaxPool}(\mathbf{a}_{i,\,p_b:p_b+r-1}),
\qquad
\hat{\mathbf{a}}_{i}
= \frac{\bar{\mathbf{a}}_{i}}
{\lVert\bar{\mathbf{a}}_{i}\rVert_1}.
\label{eq:qsa-teacher-pooling}
\end{equation}
Letting $B=\lfloor n/r\rfloor$, the pooled teacher distributions $\hat{\mathbf{a}}_i\in\mathbb{R}^{B}$ and the indexer scores $\mathbf{I}_i\in\mathbb{R}^{B}$ share the same block dimension.
The normalized teacher scores are distilled into the indexer by minimizing
\begin{equation}
\mathcal{L}_{\mathrm{KL}}
= \frac{1}{N}\sum_i
  D_{\mathrm{KL}}\left(
      \hat{\mathbf{a}}_{i,:}
      \,\middle\Vert\,
      \operatorname{Softmax}(\mathbf{I}_{i,:})
  \right),
\label{eq:qsa-distillation-loss}
\end{equation}
where $N$ is the number of query tokens. Only complete key blocks are included in the KL loss for each query.
During the warm-up stage, only the indexer is trained for 1,000 steps, using a learning rate of $1\times10^{-3}$.
Each step comprises $8$ sequences of 256K tokens, amounting to approximately 2B training tokens in total.

\item \textbf{Stage 2: Sparse Training.}
After indexer initialization, the entire backbone is trained under the guidance of the indexer to adapt to sparse attention patterns.
Following Eq.~\eqref{eq:qsa-topk-blocks}, QSA first selects the top-$K_B$ blocks.
These blocks are expanded to token indices and combined with the tail tokens in the final incomplete block:
\begin{equation}
    \mathcal{S}_i
    = \operatorname{Expand}(\mathcal{B}_i)
      \cup
      \left\{r\left\lfloor\frac{i+1}{r}\right\rfloor,\ldots,i\right\}.
    \label{eq:qsa-token-set}
\end{equation}
The $\operatorname{Expand}$ operator maps the selected blocks to their token indices. The resulting set $\mathcal{S}_i$ is used for core attention computation. In this stage, the indexer KL loss is computed only over the top-$K_B$ blocks in $\mathcal{B}_i$.
Before evaluating the KL divergence, the teacher probabilities within $\mathcal{B}_i$ are renormalized to sum to one:
\begin{equation}
    \mathcal{L}_{\mathrm{KL}}
    = \frac{1}{N}\sum_i
      D_{\mathrm{KL}}\left(
          \hat{\mathbf{a}}_{i,\mathcal{B}_i}
          \,\middle\Vert\,
          \operatorname{Softmax}(\mathbf{I}_{i,\mathcal{B}_i})
      \right).
    \label{eq:qsa-stage2-kl}
\end{equation}
During the final stage of CPT, QSA is enabled and the backbone and indexer are jointly trained for 8,000 steps using a learning rate of $2.5\times10^{-5}$. Each step comprises $96$ sequences of 256K tokens, totaling roughly 200B training tokens.

\end{itemize}

\begin{figure}[t]
    \centering
    \includegraphics[width=0.93\linewidth]{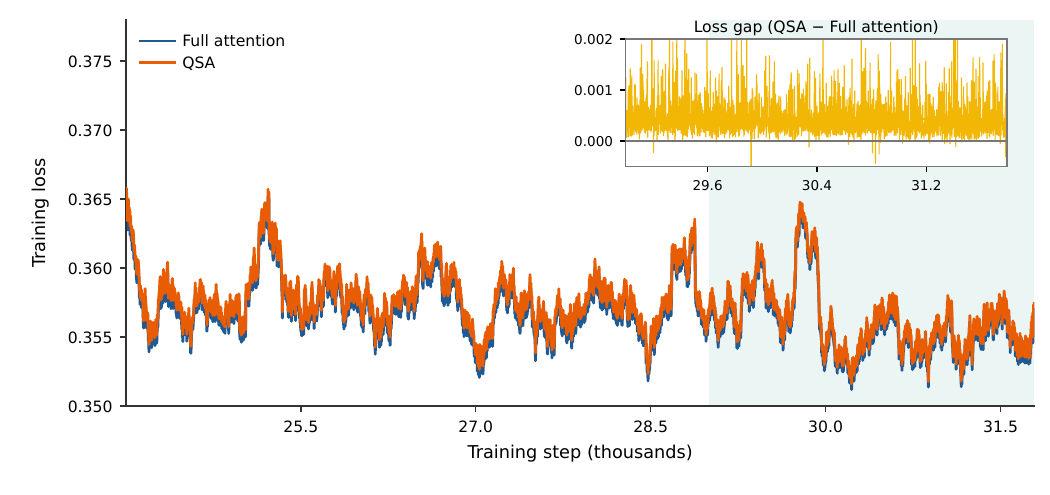}
    \caption{Training LM loss with and without QSA. Curves are smoothed with a
    200-step moving average. The shaded region marks the final stage of
    continued pretraining, and the inset shows the per-step loss difference
    between QSA and the full-attention baseline in this region.}
    \label{fig:qsa-training-loss}
\end{figure}

\paragraph{Implementation Details.}
For Qwen3.8-Flash-Next, all full-attention layers in the backbone and MTP module are replaced with QSA to improve inference efficiency on long sequences.
The compressed lightweight indexer adopts an MQA structure with four query heads and one shared key head, using the partial-RoPE configuration described above.
With a token budget of $K=2048$ and a compression ratio of $r=4$,
QSA selects up to $512$ complete blocks for each query
and additionally includes the tail tokens in the final incomplete block.
Accurate importance estimates from the indexer enable the model to closely match the LM-loss trajectory of full attention during Stage 2 sparse training.
As shown in Fig.~\ref{fig:qsa-training-loss}, the two loss curves remain highly consistent,
with the overall loss difference on the order of $10^{-4}$.

For efficient training, we implement a fused QSA kernel that jointly computes sparse attention outputs and the KL loss without materializing intermediate results,
substantially reducing memory consumption.
The compressed indexer scores the sequence after compression, reducing both indexer computation and top-$k$ selection overhead. 
In addition, multi-step MTP reuses top-$k$ indices across prediction steps to further reduce draft-model inference costs.

\paragraph{Evaluation of QSA.}
We evaluate QSA by comparing Qwen3.8-Flash-Next equipped with QSA against its full-attention baseline. Table~\ref{tab:qsa-pt-eval} reports the results across knowledge, STEM, reasoning, multilingual, and coding benchmarks,
providing a broad assessment of whether sparse attention affects the general capabilities of the model beyond long-context retrieval.
QSA matches or outperforms the full-attention baseline on seven of the eight benchmarks and improves the average score from 75.9 to 76.8.
The remaining differences are small and do not indicate systematic degradation in any task category.
Overall, these results show that QSA preserves the general capabilities of the
model on short-context tasks while enabling more efficient long-context inference.

\begin{table*}[t]
\centering
\small
\caption{Model performance comparison between Qwen3.8-Flash-Next with full attention and QSA across widely used knowledge, STEM, reasoning, multilingual, and coding benchmarks. Bold values indicate the best result for each benchmark.}
\label{tab:qsa-pt-eval}
\resizebox{\textwidth}{!}{
\begin{tabular}{@{\extracolsep{\fill}}c cc cc c c cc c@{}}
\toprule
& \multicolumn{2}{c}{Knowledge}
& \multicolumn{2}{c}{STEM}
& \multicolumn{1}{c}{Reasoning}
& \multicolumn{1}{c}{Multilingual}
& \multicolumn{2}{c}{Code}
& \\
\cmidrule(lr){2-3}
\cmidrule(lr){4-5}
\cmidrule(lr){6-6}
\cmidrule(lr){7-7}
\cmidrule(lr){8-9}
Method
& MMLU-Pro
& SuperGPQA
& MATH
& GSM8K
& BBH
& MMMLU
& EvalPlus
& MultiPL-E
& Avg. \\
\midrule
Full Attn & 72.9& 51.7& 69.8 & 91.0& 90.4
& \textbf{81.8}& 70.8& 78.4 & 75.9 \\
w/ QSA & \textbf{73.7}& \textbf{52.1}& \textbf{71.6} & \textbf{92.2}& \textbf{91.6} & 81.1& \textbf{72.3}& \textbf{79.8} & \textbf{76.8} \\
\bottomrule
\end{tabular}
}
\end{table*}

\begin{table*}[t]
\centering
\small
\caption{Long-context retrieval performance of Qwen3.8-Flash-Next on RULER and 8-needle MRCR.
RULER scores are averaged over sequence-length ranges. 
Bold values indicate the best result in each setting;
Avg. is the macro-average across the two benchmarks.}
\label{tab:qsa-long-context-eval}
\setlength{\tabcolsep}{4pt}
\begin{tabular}{l cccc cccc c}
\toprule
& \multicolumn{4}{c}{RULER}
& \multicolumn{4}{c}{MRCR}
& \\
\cmidrule(lr){2-5}
\cmidrule(lr){6-9}
Method
& $\leq$128K
& 128--256K
& 256--512K
& 512K--1M
& 128K
& 256K
& 512K
& 1M
& Avg.\\
\midrule
Full Attn
& 99.84 & \textbf{99.81} & 97.65 & 90.08
& \textbf{97.14} & \textbf{94.20} & 30.66 & 20.71 & 78.76 \\
w/ QSA
& \textbf{99.89} & 99.62 & \textbf{98.95} & \textbf{93.00}
& 95.98 & 93.00 & \textbf{40.53} & \textbf{26.44}
& \textbf{80.93} \\
\bottomrule
\end{tabular}
\end{table*}

Beyond short-context benchmarks, we evaluate the retrieval capability of QSA on two widely used long-context benchmarks, RULER~\citep{hsieh2024ruler} and MRCR~\citep{openai2025mrcr}.
We evaluate RULER at sequence lengths from 4K to 1000K and use the 8-needle MRCR setting at lengths from 128K to 1M, as reported in Table~\ref{tab:qsa-long-context-eval}.
QSA remains comparable to full attention at shorter lengths and performs better as the context grows.
In particular, QSA improves the RULER score from 90.08 to 93.00 beyond 512K. On MRCR, the score increases from 30.66 to 40.53 at 512K and from 20.71 to 26.44 at 1M.
These results suggest that QSA improves long-context inference efficiency without sacrificing task performance,
while delivering further gains at longer sequence lengths.

\begin{table*}[h]
\centering
\small
\caption{Mean MTP accepted length with full attention and QSA under four-step speculative decoding.}
\label{tab:qsa-mtp-eval}
\setlength{\tabcolsep}{4pt}
\begin{tabular}{l cccccc}
\toprule
Method & MT-Bench & GSM8K & MATH
& HumanEval & MBPP & Avg. \\
\midrule
Full Attn & 3.44 & 4.19 & 4.29 & 4.24 & 4.12 & 4.06 \\
w/ QSA & 3.47 & 4.20 & 4.30 & 4.26 & 4.13 & 4.07 \\
\bottomrule
\end{tabular}
\end{table*}
We further examine the effect of QSA on the multi-token prediction (MTP) module.
Beyond replacing the attention layers in MTP with QSA, we follow GLM~\citep{glm5team2026glm5vibecodingagentic}
and reuse the top-$k$ indices across speculative decoding steps to further improve draft-model efficiency.
To assess whether reusing QSA affects MTP performance, we conduct four-step speculative decoding experiments on benchmarks from different domains, as reported in Table~\ref{tab:qsa-mtp-eval}. 
The results show no significant change in the mean accepted length after QSA reuse.

\begin{figure}[t]
    \centering
    \begin{subfigure}[t]{0.45\linewidth}
        \centering
        \includegraphics[width=\linewidth]{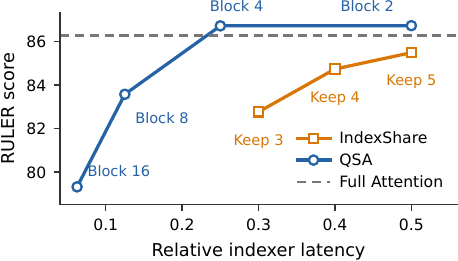}
        \caption{Compression strategies.}
        \label{fig:qsa-compression-ratio}
    \end{subfigure}
    \hfill
    \begin{subfigure}[t]{0.45\linewidth}
        \centering
        \includegraphics[width=\linewidth]{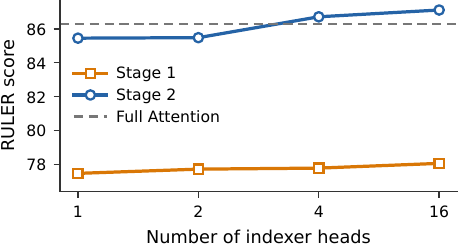}
        \caption{Indexer head configurations.}
        \label{fig:qsa-indexer-heads}
    \end{subfigure}
    \caption{Architecture ablations of QSA on RULER. 
    (a) QSA performance with different micro-block sizes;
    ``Keep $x$'' indicates the number of IndexShare indexer layers retained for computation. 
    (b) Performance with different numbers of indexer query heads after dense distillation and sparse training.}
    \label{fig:qsa-architecture-ablation}
\end{figure}

\paragraph{Architecture Ablation.}
We conduct architectural ablations of QSA at the 35B-A3B scale, focusing on the compression ratio and the number of indexer heads.
As shown in Fig.~\ref{fig:qsa-architecture-ablation}, we use RULER~\citep{hsieh2024ruler} scores at sequence lengths up to 1M as the evaluation metric.
All experiments are evaluated using the CPT models obtained after stage 2 sparse training.

For the compression-ratio ablation, we evaluate QSA with different micro-block sizes.
We additionally include training-aware IndexShare~\citep{bai2026indexcache} as a baseline, 
which reduces indexing overhead by uniformly sharing top-$K$ indices across adjacent full-attention layers in the hybrid model.
In Fig.~\ref{fig:qsa-architecture-ablation}(a), we report the RULER performance of both approaches against their estimated reductions in indexer latency.
QSA matches the full-attention baseline at a relative indexer latency of 0.25, whereas IndexShare remains below the baseline at 0.5.
For IndexShare, 0.5 denotes sharing a single index across two full-attention layers separated by three GDN layers.
This result highlights the advantage of intra-layer compression in hybrid architectures, where cross-layer index sharing can be limited by low inter-layer similarity.

The number of query heads directly affects indexer efficiency, particularly during the prefill stage.
We therefore evaluate MQA indexers with varying numbers of query heads on RULER, as shown in Fig.~\ref{fig:qsa-architecture-ablation}(b).
After dense initialization, directly applying the indexer for sparse attention leads to a clear performance drop. 
A brief period of joint training allows the backbone to adapt to the sparse attention pattern and recover to the full-attention level.
The results show that QSA maintains performance with only a small number of indexer query heads,
far fewer than used in the core attention module.
To balance inference speed and accuracy,
we ultimately adopt 4 query heads as a lightweight indexer configuration.

\begin{figure}[t]
    \centering
    \includegraphics[width=\linewidth]{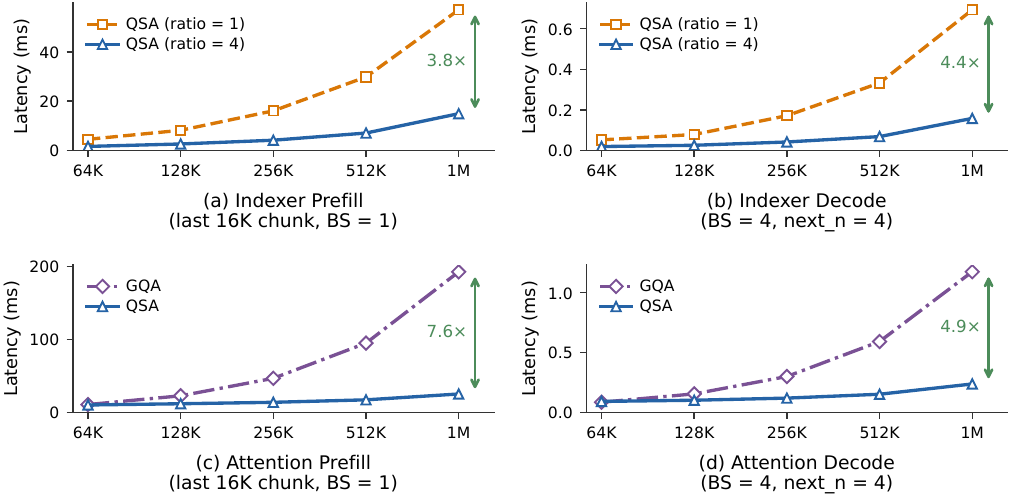}
    \caption{Kernel-level latency of QSA across context lengths during prefill
    and decode. Panels
    (a,b) compare indexer latency under different compression ratios, while
    panels (c,d) compare kernel-level attention latency between dense GQA and
    QSA, including both the indexer and sparse core attention. Chunked prefill
    uses a 16K-token chunk with batch size 1; decode uses batch size 4 and
    \texttt{next\_n}$=4$, corresponding to three MTP prediction steps. Arrows
    indicate speedups at a context length of 1M.}
    \label{fig:qsa-efficiency}
\end{figure}

\paragraph{Efficiency Analysis.}
By compressing the key sequence with a ratio of $r$, QSA reduces indexer complexity from $O(n^2)$ to $O(n^2/r)$, which yields substantial efficiency gains on long sequences.
For the indexer, sequence compression directly reduces the computation of MQA logits and top-$k$ selection,
yielding a speedup comparable to the compression ratio.
As shown in Fig.~\ref{fig:qsa-efficiency}(a,b), this substantially reduces inference costs on long sequences, where the indexer becomes a major bottleneck.

We further evaluate the kernel-level speedup of QSA, accounting for both the indexer cost and sparse core attention computation.
For the dense-attention baseline, we use the paged GQA implementation provided by FlashInfer~\citep{ye2025flashinfer}.
Prefill is evaluated under a chunked-prefill setting, while decode includes three additional MTP steps.
As shown in Fig.~\ref{fig:qsa-efficiency}(c,d), QSA provides speedups from a context length of 64K, with increasingly larger gains as the sequence length grows.
At a context length of 1M, QSA achieves $7.6\times$ and $4.9\times$ attention-module speedups for prefill and decode, respectively, demonstrating its scalability for long-context inference.

\subsection{Residual}
\label{sec:residual}

\paragraph{Motivation.}
\label{sec:residual-motivation}
Residual connections give every block a direct path to the network output~\citep{he2016resnet}. 
Pre-normalization keeps training stable at scale~\citep{xiong2020layernorm}, but it attenuates the signal each layer receives: every block reads the same stream, so a feature written early must compete with everything written after it. 
Several lines of work address this by adding paths that bypass the bottleneck, including dense inter-layer connectivity~\citep{huang2017densenet}, an extra residual path for attention values~\citep{zhou2024valueresidual}, and cross-layer reuse of cached states~\citep{sun2024yoco}.

Work that modifies the residual path directly falls into two families. 
The first makes each layer's read and write more expressive, following highway networks~\citep{srivastava2015highway}. 
The second widens the stream itself: Alternating Updates (AltUp)~\citep{baykal2023altup} and Hyper-Connections (HC)~\citep{zhu2024hyper} replace the single residual vector with several parallel branches. 
The two are complementary: widening adds capacity, and a richer read/write mechanism decides how that capacity is spent.

\paragraph{Widening the Residual Stream.}
\label{sec:residual-width}
We first study how much of the reported gain comes from width alone, using a
simplified variant of AltUp~\citep{baykal2023altup} that fits a pre-norm
network. The residual state before block $\ell$ is a set of $n_r$ branches
$\bm{R}^{(\ell)}\in\mathbb{R}^{n_r\times d}$, where $d$ is the hidden size
and $\bm{R}^{(\ell)}_i$ denotes branch $i$. Each block holds $n_r$ learnable
scalars $\bm{h}\in\mathbb{R}^{n_r}$ and reads its input as a weighted sum of
the branches,
\begin{equation}
\bm{x}^{(\ell)} = \sum_{i=1}^{n_r} h_i\,\bm{R}^{(\ell)}_i .
\label{eq:altup-read}
\end{equation}
The block output $\bm{y}^{(\ell)}$ is then written back to a single branch,
chosen round-robin by depth:
\begin{equation}
\bm{R}^{(\ell+1)}_i =
\bm{R}^{(\ell)}_i + \mathbf{1}\left[i = \ell \bmod n_r\right]\bm{y}^{(\ell)} .
\label{eq:altup-write}
\end{equation}
This adds $n_r$ parameters per block and no matrix multiplication, so its
compute cost is negligible; the extra cost is the memory traffic of carrying
$n_r$ branches instead of one. Even so, it lowers the training loss of a
25B-A3B MoE model trained on 400B tokens by roughly $0.01$. Widening alone
therefore yields a substantial loss reduction. The question that remains is
how much read/write machinery is worth adding on top of the widened stream to
reach a good balance of performance, efficiency, and stability.

\paragraph{Hyper-Connections (HC).}
\label{sec:residual-hc}
HC generalizes Eq.~\eqref{eq:altup-read} and Eq.~\eqref{eq:altup-write} into three learnable operators.
A read operator $\bm{H}_{\mathrm{mix}}$ forms the block input, a write operator $\bm{H}_{\mathrm{combine}}$ distributes the block output over branches, and a mixing operator $\bm{H}_{\mathrm{res}}$ exchanges information between branches:
\begin{align}
\bm{x}^{(\ell)} &= \bm{H}_{\mathrm{mix}}^{\top}\bm{R}^{(\ell)}, \label{eq:hc-mix}\\
\bm{y}^{(\ell)} &= \mathcal{F}^{(\ell)}\left(\operatorname{Norm}\left(\bm{x}^{(\ell)}\right)\right), \label{eq:hc-block}\\
\bm{R}^{(\ell+1)} &= \bm{H}_{\mathrm{res}}\,\bm{R}^{(\ell)}
   + \bm{H}_{\mathrm{combine}}\,\bm{y}^{(\ell)\top}, \label{eq:hc-combine}
\end{align}
with $\bm{H}_{\mathrm{mix}},\bm{H}_{\mathrm{combine}}\in\mathbb{R}^{n_r}$ and $\bm{H}_{\mathrm{res}}\in\mathbb{R}^{n_r\times n_r}$. In the notation of HC these are $\bm{A}_m$, $\bm{B}$ and $\bm{A}_r$.
The three operators are predicted from the residual state. Let $\overline{\bm{R}}=\operatorname{norm}(\bm{R}^{(\ell)})$ be a normalized view of the branches. Each operator is then the sum of a static term and a data-dependent term:
\begin{align}
\bm{H}_{\mathrm{mix}} &= \bm{H}^{\mathrm{s}}_{\mathrm{mix}}
   + \bm{\lambda}_{m}\odot\phi\left(\overline{\bm{R}}\,\bm{W}_{m}\right),
   \label{eq:hc-param-mix}\\
\bm{H}_{\mathrm{combine}} &= \bm{H}^{\mathrm{s}}_{\mathrm{combine}}
   + \bm{\lambda}_{c}\odot\phi\left(\overline{\bm{R}}\,\bm{W}_{c}\right),
   \label{eq:hc-param-combine}\\
\bm{H}_{\mathrm{res}} &= \bm{H}^{\mathrm{s}}_{\mathrm{res}}
   + \bm{\lambda}_{r}\odot\phi\left(\overline{\bm{R}}\,\bm{W}_{r}\right),
   \label{eq:hc-param-res}
\end{align}
where $\bm{H}^{\mathrm{s}}_{\star}$ is a learnable static term, $\bm{W}_{\star}$ projects the normalized residual, $\phi$ is an activation function, and $\bm{\lambda}_{\star}$ is a learnable scale, with $\star$ standing for any of the three operators. HC uses $\phi=\tanh$ with $\bm{\lambda}_{\star}$ initialized to $0.01$; mHC~\citep{xie2025mhc} uses a sigmoid and additionally constrains $\bm{H}_{\mathrm{res}}$ to a manifold of doubly stochastic matrices.
Specifically, we set the static terms to $\bm{H}^{\mathrm{s}}_{\mathrm{mix}} = \bm{e}_{\ell\bmod n_r}$, $\bm{H}^{\mathrm{s}}_{\mathrm{combine}} = \bm{1}$, and $\bm{H}^{\mathrm{s}}_{\mathrm{res}} = \bm{I}$. With $\bm{W}_{\star}=\bm{0}$, this initialization starts the widened network exactly at the pre-norm baseline; with $\bm{\lambda}_{\star}=\bm{0}$ throughout, the operators stay static and the widened stream costs no extra computation, which recovers the simplified AltUp variant above.

\paragraph{Design Ablation.}
\label{sec:residual-ablation}
Starting from the static operators, we kept the added expressiveness only where
it paid for itself.
Tab.~\ref{tab:residual-ablation} reports the endpoints of that progression,
evaluated with the benchmark suite and evaluation pipeline of
\S\ref{sec:gdn-ablation}. Five observations shaped the final design.
\begin{itemize}[leftmargin=1.5em]
\item \textbf{Bounded positive gates.} A sigmoid gate is better than $\tanh$
  in both loss and training stability. This agrees with
  mHC~\citep{xie2025mhc} and is consistent with the observation in the GDN
  and attention components that sigmoid gates outperform SiLU or $\tanh$.
\item \textbf{Data dependence.} Making $\bm{H}_{\mathrm{mix}}$ and $\bm{H}_{\mathrm{combine}}$ data-dependent reduces loss by 0.002 over the static variant, whereas the static variant reduced loss by 0.021 over the pre-norm baseline. 
The benchmark gain reverses this ratio: 1.98 points from static to dynamic against 1.58 from baseline to static (Tab.~\ref{tab:residual-ablation}).  
This is one of several places in this report where loss and downstream accuracy do not move together.
\item \textbf{Read granularity matters more than write granularity.} Refining
  $\bm{H}_{\mathrm{mix}}$ in Eq.~\eqref{eq:hc-mix} from one scalar per branch
  to one weight per branch and channel helps. The same refinement of
  $\bm{H}_{\mathrm{combine}}$ in Eq.~\eqref{eq:hc-combine} gives almost
  nothing, so the write stays a per-branch scalar.
\item \textbf{Read all branches.} Predicting the operators from all branches
  is better than using only the last branch or pooling the branches first.
  Normalizing each branch separately, that is, a group RMSNorm over the
  widened stream, gives a further gain.
\item \textbf{$\bm{H}_{\mathrm{res}}$ adds little.} Once the read and the
  write are expressive enough, adding the $n_r\times n_r$ mixing operator
  brings no significant improvement.
\end{itemize}

\begin{table*}[t]
\centering
\small
\setlength{\tabcolsep}{3.5pt}
\caption{Residual read/write ablation on 25B-A3B MoE models
 trained on 560B tokens. The benchmarks and the evaluation pipeline are those
 of \S\ref{sec:gdn-ablation}. All widened variants use $n_r=4$ branches;
 \emph{static} is the case $\bm{\lambda}_{\star}=\bm{0}$ of
 Eq.~\eqref{eq:hc-param-mix}--\eqref{eq:hc-param-res}, \emph{dynamic} its
 data-dependent counterpart.}
\label{tab:residual-ablation}
\resizebox{\textwidth}{!}{%
\begin{tabular}{lrrrrrrrrrrr}
\toprule
& & \multicolumn{3}{c}{Knowledge} & \multicolumn{2}{c}{STEM} & Reasoning
  & Multilingual & \multicolumn{2}{c}{Code} & \\
\cmidrule(lr){3-5} \cmidrule(lr){6-7} \cmidrule(lr){8-8}
  \cmidrule(lr){9-9} \cmidrule(lr){10-11}
Residual & Loss & MMLU & MMLU-Pro & SuperGPQA & MATH & GSM8K & BBH & MMMLU
  & EvalPlus & MultiPL-E & Avg. \\
\midrule
Pre-norm
& 1.617 & 64.29 & 38.40 & 21.78 & 53.92 & 77.41 & 64.73 & 51.26 & 49.25
  & 37.15 & 50.91 \\
mHC (static)
& 1.596 & 64.62 & 43.69 & 22.20 & 55.08 & 78.05 & 65.42 & 52.78 & 49.59
  & 40.94 & 52.49 \\
mHC (dynamic)
& 1.594 & 66.11 & 45.84 & \textbf{24.20} & 59.54 & \textbf{78.51} & 66.01
  & \textbf{56.61} & \textbf{52.16} & 41.30 & 54.47 \\
GR
& 1.590 & \textbf{66.69} & \textbf{46.02} & 23.80 & \textbf{61.18} & 78.20
  & \textbf{66.54} & 56.19 & 51.36 & \textbf{42.00} & \textbf{54.66} \\
\bottomrule
\end{tabular}
}
\end{table*}

Widening the stream with static operators is already worth $1.58$ points of average accuracy over the pre-norm baseline, and making the read and the
write data-dependent adds a further $1.98$. The loss gap between mHC static and dynamic is only $0.002$, yet the benchmark gap is substantial; this is a case where loss alone would have understated the value of the change,
and it reinforced our practice of checking benchmarks alongside loss
throughout the design process. We describe the concrete design of GR below.

\paragraph{Gated Residual.}
\label{sec:residual-gr}
In separate work~\citep{qiu2026unified}, we had found that adding a lightweight elementwise self-gate
after RMSNorm, which we call GatedNorm, markedly improves training stability:
\begin{equation}
\operatorname{GatedNorm}(\bm{u}) = \operatorname{RMSNorm}(\bm{u})\odot
\sigma\left(\bm{W}_{2}\operatorname{SiLU}\left(\bm{W}_{1}
\operatorname{RMSNorm}(\bm{u})\right)\right),
\label{eq:gatednorm}
\end{equation}
where $\bm{W}_1$ and $\bm{W}_2$ form a low-rank bottleneck.
The read the ablation arrives at, elementwise and data-dependent with a
sigmoid gate, is exactly Eq.~\eqref{eq:gatednorm} applied to the widened
stream. We therefore merge the two into a single operator and call the result
Gated Residual (GR).

GR first normalizes each branch independently,
\begin{equation}
\widehat{\bm{R}}_i = \operatorname{RMSNorm}\left(\bm{R}_i;\bm{\gamma}_i\right),
\qquad i=1,\dots,n_r,
\label{eq:gr-norm}
\end{equation}
each with its own gain $\bm{\gamma}_i\in\mathbb{R}^{d}$. It then predicts 
elementwise gating scores per branch and channel from all branches, and averages the gated
branches into the block input:
\begin{align}
\bm{G} &= \operatorname{unvec}\,\sigma\left(\bm{W}_{u}
\operatorname{SiLU}\left(
   \tfrac{1}{n_r}\bm{W}_{d}\operatorname{vec}(\widehat{\bm{R}})\right)\right)
   \in\mathbb{R}^{n_r\times d},
   \label{eq:gr-gate}\\
\bm{x} &= \frac{1}{n_r}\sum_{i=1}^{n_r}\bm{G}_i\odot\widehat{\bm{R}}_i ,
   \label{eq:gr-read}
\end{align}
where $\operatorname{vec}$ stacks the branches into one vector of length
$n_r d$ and $\operatorname{unvec}$ is its inverse, with
$\bm{W}_{d}\in\mathbb{R}^{r\times n_r d}$ and
$\bm{W}_{u}\in\mathbb{R}^{n_r d\times r}$ for a bottleneck rank $r = d / 8$.
The block output $\bm{y}=\mathcal{F}(\bm{x})$ is written to every branch
through one data-dependent scalar per branch:
\begin{align}
\bm{s} &= 2\,\sigma\left(\tfrac{1}{n_r}\bm{W}_{w}
\operatorname{vec}(\widehat{\bm{R}})\right)
   \in\mathbb{R}^{n_r},
   \label{eq:gr-write-gate}\\
\bm{R}'_i &= \bm{R}_i + s_i\,\bm{y} ,
   \label{eq:gr-write}
\end{align}
with $\bm{W}_{w}\in\mathbb{R}^{n_r\times n_r d}$.
Eq.~\eqref{eq:gr-gate} and Eq.~\eqref{eq:gr-write-gate} are the read and
write operators of Eq.~\eqref{eq:hc-param-mix} and
Eq.~\eqref{eq:hc-param-combine} with $\phi=\sigma$, an elementwise
$\bm{H}_{\mathrm{mix}}$, a per-branch scalar $\bm{H}_{\mathrm{combine}}$,
and $\overline{\bm{R}}$ the group-RMSNorm of all branches.
The static term $\bm{H}^{\mathrm{s}}_{\star}$ brings no improvement for GR.
With the current configuration, no special initialization of the learnable
weights is needed, and the static term contributes negligibly; standard
random initialization, as used in the backbone, suffices.
We use $n_r=4$ branches, with a separate GR module for the attention block
and the MLP block of every layer.

In Tab.~\ref{tab:residual-ablation}, GR and mHC (dynamic) differ mainly in
the elementwise $\bm{H}_{\mathrm{mix}}$ and the removal of
$\bm{H}_{\mathrm{res}}$; at this scale the two perform comparably. The
efficiency advantage of GR is that removing $\bm{H}_{\mathrm{res}}$
eliminates a full read of the residual state per block, reducing memory
traffic. 
The stability advantage is twofold: GatedNorm itself improves
training stability (analysed in \S\ref{sec:stability}), and dropping $\bm{H}_{\mathrm{res}}$, which requires separate constraints, removes a potential source of instability.

GR belongs to the same family as HC, mHC and VWN~\citep{seed2025vwn}; what
differs is where the extra expressiveness is spent. HC and mHC keep the read
and the write as per-branch scalars and put capacity into
$\bm{H}_{\mathrm{res}}$, which mHC further constrains to be doubly
stochastic. VWN keeps those scalars as well and instead widens the token
embedding, splitting it into many narrow segments; this splitting likewise
pursues finer-grained read and write operations. GR spends its expressiveness
on the read: the gate is elementwise, and $\bm{H}_{\mathrm{res}}$ is dropped
altogether, which the ablation shows costs nothing and which removes a read
of the whole residual state per block, the dominant inference cost of a
widened stream.

Because the read in Eq.~\eqref{eq:gr-read} already normalizes and gates, GR
also replaces the block's pre-normalization rather than sitting in front of it: Eq.~\eqref{eq:hc-block} loses its $\operatorname{Norm}$, and widening adds no normalization layer. 
And with no mixing operator the branches stay separate, since a branch is only ever written by blocks and read through Eq.~\eqref{eq:gr-read}. 
This makes the information flow easy to follow, which we use in the analysis in \S\ref{sec:residual-analysis}.

\paragraph{Comparison with Attention Residual.}
\label{sec:residual-attnres}

\begin{wraptable}{r}{0.48\linewidth}
\centering
\small
\vskip -0.15in
\setlength{\tabcolsep}{4pt}
\caption{Residual designs at $28$ layers, with and without GatedNorm (GN). 
Loss is the final training loss; $S$ is the number of sublayers summed into one Block AttnRes representation. Subscripts give the change from the column on the left.}
\vskip -0.05in
\label{tab:residual-attnres}
\begin{tabular}{lcc}
\toprule
Residual design & Loss & \makecell{Loss + GN} \\
\midrule
Pre-norm residual    & 1.789 & $1.787_{-0.002}$ \\
Block AttnRes, $S=4$ & 1.773 & $1.768_{-0.005}$ \\
Block AttnRes, $S=2$ & 1.770 & $1.766_{-0.004}$ \\
Full AttnRes         & 1.762 & $\mathbf{1.758}_{-0.004}$ \\
GR ($n_r=4$)         & --- & 1.762 \\
\bottomrule
\end{tabular}
\vskip -0.15in
\end{wraptable}

Attention Residual
(AttnRes)~\citep{kimiteam2026attentionresiduals} uses a softmax attention over earlier layers' outputs to determine each sublayer's read. Full AttnRes attends over the output of every preceding sublayer. 
Block AttnRes partitions the $L$
sublayers into blocks of $S$, sums each block into one representation, and attends over those.

Tab.~\ref{tab:residual-attnres} compares both variants against GR in our setting, a $28$-layer model ($L=56$ sublayers), each with and without GN. 
Full AttnRes is the strongest setting of that family and lands level with GR at $1.762$, while summarizing blocks costs $0.008$ at $S=2$ and $0.011$ at $S=4$. 
The same ordering holds deeper: at $48$ layers, Block
AttnRes at $S=4$ reaches $1.711$ against $1.707$ for GR.
GN lowers the loss at every setting, by $0.004$--$0.005$ on AttnRes and by $0.002$ on the plain pre-norm baseline. 
The gate helps more when the input a sublayer reads is more complex, which is the same gate GR carries inside its read.

\paragraph{What the Branches Are Used For.}
\label{sec:residual-analysis}
The ablation shows that GR improves both loss and benchmarks, but what mechanism produces the gain? 
As GR has no residual branch mixing, each branch is
a plain accumulator of past outputs, and we can decompose exactly what each block reads into contributions from every earlier block. 
We use this decomposition to compare a GR model against an otherwise identical reference without GR, and find that the extra width is spent on several specific paths: one branch preserves early attention outputs across many layers, while the other three stay local.

\emph{The statistic.}
Branch $c$ before block $v$ holds
\begin{equation}
\bm{R}^{(v)}_c = \bm{R}^{(0)}_c + \sum_{u<v} s^{(u)}_c\,\bm{y}^{(u)},
\label{eq:gr-accumulator}
\end{equation}
where $\bm{R}^{(0)}_c$ is the initial value of branch $c$ (the token
embedding), $\bm{y}^{(u)}$ is the output of block $u$, and $s^{(u)}_c$ is
the scalar write gate of Eq.~\eqref{eq:gr-write-gate} controlling how much
of $\bm{y}^{(u)}$ enters branch $c$.

Block $v$ reads this branch through the gated read of
Eq.~\eqref{eq:gr-read}: it normalizes the branch, scales it by a learned
per-channel gain $\bm{\gamma}_c$ (from Eq.~\eqref{eq:gr-norm}), gates the
result elementwise with $\bm{G}^{(v)}_c$ (the data-dependent gate of
Eq.~\eqref{eq:gr-gate}), and averages over branches. Since normalization
divides by $\operatorname{rms}(\bm{R}^{(v)}_c)$, the contribution of block
$u$ to block $v$'s input is
\begin{equation}
\bm{a}_{u\to v}
= \frac{1}{n_r}\sum_{c=1}^{n_r}
  \bm{G}^{(v)}_c\odot\bm{\gamma}_c\odot
  \frac{s^{(u)}_c\,\bm{y}^{(u)}}{\operatorname{rms}\left(\bm{R}^{(v)}_c\right)},
\label{eq:gr-attribution}
\end{equation}
evaluated at the gate values the forward pass actually took. Each factor has
a direct reading: $s^{(u)}_c\,\bm{y}^{(u)}$ is what block $u$ deposited on
branch $c$; the division by $\operatorname{rms}(\bm{R}^{(v)}_c)$ accounts
for the normalization applied at read time; $\bm{\gamma}_c$ rescales each
channel; and $\bm{G}^{(v)}_c$ decides how much of this branch block $v$
actually uses.

\begin{figure}[t]
  \centering
  \includegraphics[pagebox=cropbox,width=\linewidth]{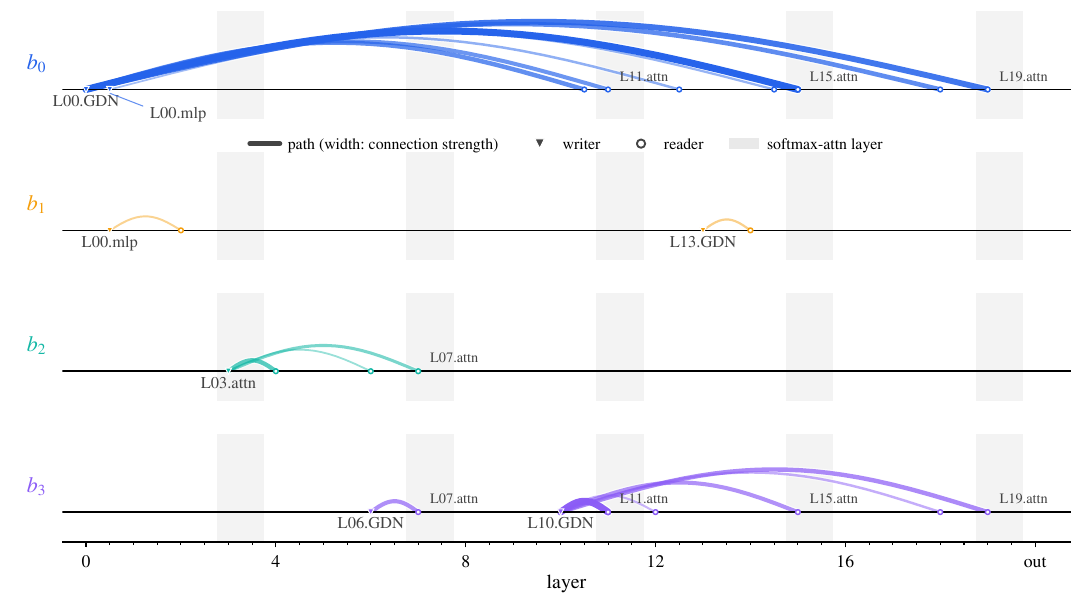}
  \vskip -0.1in
\caption{\textbf{Cross-layer paths added by GR.} Each row corresponds to one residual branch; a connection runs from the sublayer that wrote
into that branch to a later sublayer that reads it back. Vertical position encodes the number of layers skipped; line width and opacity encode $\Delta_{uv}$
(Eq.~\eqref{eq:gr-share}), the additional share of the reader's input supplied by this writer compared to a single residual stream. 
Shaded regions denote softmax-attention layers, every fourth in this hybrid; the rest are GDN, and
sublayers are named accordingly (\texttt{L00.GDN}, \texttt{L03.attn}, \texttt{L00.mlp}). 
Readers are named where they are softmax-attention sublayers, which is where the long-range paths land. 
One branch carries long-range paths (all connections on $b_0$ originate at layer 0 and land past layer 10), while the other
three stay local (median skips of 1.2--3.5 layers). The extra width is spent on a small number of specific paths, mostly preserving early GDN output across depth and
delivering it to the softmax-attention layers. Thresholds and counts are given in the text.}
  \label{fig:gr-excess-paths}
\end{figure}

We report the normalized magnitude of this contribution,
\begin{equation}
\pi_{uv} = \frac{\lVert\bm{a}_{u\to v}\rVert}
                {\sum_{u'<v}\lVert\bm{a}_{u'\to v}\rVert},
\label{eq:gr-share}
\end{equation}
the fraction of block $v$'s input that came from block $u$. Because
Eq.~\eqref{eq:gr-attribution} splits each contribution by branch before
summing, we can see which branch carries each path. The decomposition is
exact: every reader's shares sum to one to within $3\times10^{-8}$.

We compare a 20-layer MoE trained with GR against an otherwise identical reference without GR, same recipe, data, optimizer and training step, probed on the same tokens. 
We report the difference
$\Delta_{uv}=\pi^{\mathrm{GR}}_{uv}-\pi^{\mathrm{ref}}_{uv}$, since
subtracting the reference removes the pattern common to all residual networks,
namely that nearby writers dominate.

\emph{What the figure shows.}
Fig.~\ref{fig:gr-excess-paths} shows the $21$ paths (out of $780$ ordered
pairs) with $\Delta_{uv}\ge 0.05$ over a skip of at least one layer. To
calibrate: at layer $15$ there are $30$ writers, so an equal split gives
each about $0.03$; a share of $0.13$ is four times that.

One branch carries long-range paths while the other three stay local. Which
branch it is does not matter, since the branches are exchangeable at
initialization, but the pattern is consistent: across five GR checkpoints,
each has exactly one such branch, with a typical skip of $10.9$ layers
against $3.4$--$3.9$ for the rest. Three examples illustrate the pattern;
each gives the share under the reference model (a single residual stream
without GR) and then under GR.
\begin{itemize}[leftmargin=1.5em]
\item Layer $0$ GDN $\to$ layer $15$ attention: the share rises from $0.020$ in the reference to $0.138$ in GR. 
    This share holds at $0.072$--$0.138$ across every reader
  from layer $10$ to $19$ with no downward trend.
  This confirms the outsized role of the first layer in preserving information across depth, consistent with prior observations~\citep{elhage2021mathematical,men2024shortgpt}.
\item Layer $10$ GDN $\to$ layer $11$ attention:
  $\Delta_{uv}=0.117$. The gain is as large as the path above, but over a single layer: GR strengthens short-range connections as well.
\item Layer $0$ MLP, on two branches at once: to layer $15$ on the long-range branch, where its share rises from $0.008$ in the reference no-GR model to $0.058$ in GR; and to layer $2$ on a local one, where it rises from $0.139$ to $0.192$.
The same output reaches a nearby and a distant reader with different strengths, which a single stream cannot express since it has only one decay rate for every writer.
\end{itemize}

The typical long-range path in branch 0 follows from Eq.~\eqref{eq:gr-accumulator}: the branch is written most heavily at layer 0 and barely updated thereafter, so that layer-0 information remain accessible to all subsequent layers.
Furthermore, the sublayers that read most heavily from the GR branches are predominantly the softmax attention layers, indicating that global attention acts as a critical hub for integrating explicit long-range historical context that the GDN layers compress away.

To see the overall pattern, we group all $780$ paths by how many layers they skip and sum $\Delta_{uv}$ within each group, where $\Delta_{uv}$ is the extra share that GR gives to path $(u,v)$ over the reference. 
The result: adjacent-layer paths (skip $1$) collectively gain $0.96$ in share; long-range paths (skip $>12$) collectively gain $0.91$; and mid-range paths
(skip $2$--$12$) collectively lose $3.21$. The weighted-average skip is almost unchanged ($3.97$ vs $3.91$), so the total amount of cross-layer information is similar; what changes is its distribution. 
GR selects a few paths and amplifies them, at the expense of mid-range ones.

\paragraph{Inference Efficiency.}
\label{sec:residual-efficiency}
The inference cost of GR is dominated by the memory traffic of the widened residual state. We therefore looked for ways to move fewer bytes.

The first attempt was to sparsify the read. We observed that in trained models, the write at each GR layer is typically dominated by two branches.
We therefore tried introducing sparse writes, either from scratch or mid-training, where each block reads only the two branches with the highest gate values instead of all $n_r$. 
Pre-training loss and benchmarks were almost unaffected, but the quality degraded clearly after post-training, so we did not adopt it. 
More complex variants, such as varying the sparsity level across layers, did not resolve the issue. 
We note \textit{this as a case where pre-training metrics alone would have led to the wrong decision}. 
Recent work xHC~\citep{zhang2026xhc} explores using a larger $n_r$ to make sparse branch updates easier, but given the memory overhead of a larger $n_r$, we did not explore this direction further.

The second attempt was to keep the residual state in FP8. The gates in GR, gated attention, and GDN all bound the magnitude of what is written into the stream, so residual values stay in a narrow range and are well matched to a low-precision format. 
Storing the branches in FP8 halves the bytes moved for the residual state relative to BF16, with almost no loss in quality.
Finally, the read of Eq.~\eqref{eq:gr-norm}--\eqref{eq:gr-read} and the write of Eq.~\eqref{eq:gr-write-gate}--\eqref{eq:gr-write} are each fused into a single kernel, with the group RMSNorm folded into the read, so the
widened stream is traversed once per block in each direction.

\subsection{N-gram Embedding}
\label{sec:ngram-emb}
\paragraph{Motivation.}
Embedding-based memory offers a complementary dimension for scaling model capacity~\citep{Jordan2024,google2025gemma3n,rwkvcommunity2025deepembed,gemmateam2026gemma4,sadhukhan2026stem,tseng2026l3}.
N-gram embeddings further generalize unigram lookup by conditioning memory retrieval on local context rather than token identity alone~\citep{huang2021lookuplm,roy2022ngrammer,huang2025overtokenized,yu2025scone,cheng2026engram, liu2026scalingembeddings,chen2026xgram}. 
Concretely, short n-grams ending at each token serve as keys into embedding tables, and the retrieved vectors augment the corresponding token representation.
N-gram memory scales capacity with negligible additional per-token FLOPs, while deterministic addressing enables host-memory offloading and asynchronous prefetching.
In this section, we systematically study the key architectural choices for N-gram embedding.
We use multi-head hashing for embedding lookup~\citep{roy2022ngrammer,huang2025overtokenized}. 
The embeddings are further injected into residual stream through contextual gating mechanism~\citep{cheng2026engram}.
We use 300 tokens per active parameter (TPP) throughout experiments.

\subsubsection{Placement of the N-gram Embedding Layers}
\begin{table*}[h]
\centering
\caption{Effect of N-gram embedding layer placement. The total number of N-gram embedding parameters
is fixed across all settings.}
\label{tab:ngram-layer-index}
\resizebox{\textwidth}{!}{
\begin{tabular}{l c ccc cc c c cc c}
\toprule
&
& \multicolumn{3}{c}{Knowledge}
& \multicolumn{2}{c}{STEM}
& \multicolumn{1}{c}{Reasoning}
& \multicolumn{1}{c}{Multilingual}
& \multicolumn{2}{c}{Code}
&  \\
\cmidrule(lr){3-5}
\cmidrule(lr){6-7}
\cmidrule(lr){8-8}
\cmidrule(lr){9-9}
\cmidrule(lr){10-11}

Layer Index & Loss
& MMLU
& MMLU-Pro
& SuperGPQA
& MATH
& GSM8K
& BBH
& MMMLU
& EvalPlus
& MultiPL-E
& Avg. \\
\midrule
w/o N-gram emb.
& 1.585 & 62.78 & 33.43 & 20.97 & 32.52 & 59.21 & 53.40 & 54.06 & 52.13 & 45.42 & 45.44 \\
1st
& 1.541 & 64.19 & 35.25 & 21.30 & 36.20 & \bf{65.73} & 56.00 & 56.16 & 50.95 & \bf{47.45} & 47.30 \\
2nd
& 1.541 & 64.71 & \bf{35.80} & 21.49 & 37.32 & 64.00 & 57.56 & \bf{56.64} & \bf{53.09} & 45.56 & 47.94 \\
3rd
& 1.543 & 63.20 & 34.93 & 20.67 & 35.74 & 63.15 & 56.71 & 55.02 & 50.15 & 44.90 & 46.76 \\
4th
& 1.544 & 63.36 & 34.81 & \bf{22.33} & 36.20 & 61.26 & 55.32 & 55.79 & 51.05 & 45.60 & 46.89 \\
10th
& 1.544 & 64.22 & 34.99 & 20.81 & 33.80 & 62.17 & 55.54 & 54.81 & 52.69 & 43.61 & 46.62 \\
15th
& 1.543 & \bf{65.07} & 34.95 & 21.35 & 36.12 & 63.50 & 57.15 & 55.98 & 52.21 & 44.42 & 47.37 \\
25th
& 1.541 & 64.70 & 35.15 & 22.13 & 36.26 & 63.31 & 55.73 & 55.99 & 52.66 & 44.33 & 47.40 \\
\midrule
2nd + 15th
& 1.541 & 63.82 & 35.63 & 21.25 & \bf{37.48} & 63.23 & 56.52 & 55.68 & 50.44 & 43.85 & 47.01 \\
2nd + 25th
& \bf{1.540} & 64.94 & 35.40 & 21.80 & 37.40 & 64.33 & \bf{57.79} & 56.45 & 51.69 & 44.73 & 47.75 \\
\bottomrule
\end{tabular}
}
\end{table*}
We ablate the placement and number of N-gram embedding layers under a fixed parameter budget.
Single-layer variants span shallow (Layers 1-4), intermediate (Layers 10 and 15), and deep (Layer 25) locations.
For multi-layer configurations, we combine a shallow layer (Layer 2) with either an intermediate layer (Layer 15) or a deep layer (Layer 25).
The results are shown in Table~\ref{tab:ngram-layer-index}.

No single depth regime consistently dominates. 
The first two layers perform strongly, while intermediate and deep placements remain competitive.
Distributing the same parameter budget across multiple layers yields no consistent benefit. 
The marginal loss reduction from combining Layers 2 and 25 does not translate into improved downstream performance. 
A single N-gram embedding layer is sufficient. 
Moreover, the relative performance of different placements is similar under full attention and GDN (\S\ref{sec:gdn-hybrid}), suggesting that placement choice is largely insensitive to the attention mechanism.
We place it at Layer 2, \textit{allowing host-memory prefetching to overlap with the computation of the first layer.}

\subsubsection{N-gram Vocabulary Size}
We further explore the effect of scaling the vocabulary size of N-gram embeddings.

\begin{table*}[h]
\centering
\caption{Effect of N-gram vocabulary scaling under a fixed total model parameter budget. Vocabulary scales are measured relative to the base tokenizer vocabulary size (250K). The number of MoE experts is adjusted to offset the additional N-gram embedding parameters.}
\label{tab:ngram-vocab-size-samebudget}
\resizebox{\textwidth}{!}{
\begin{tabular}{l cc ccc cc c cc c}
\toprule
Vocab. Scale
& \multirow{2}{*}{Loss}
& Uncheatable
& \multicolumn{3}{c}{Knowledge}
& \multicolumn{2}{c}{STEM}
& \multicolumn{1}{c}{Reasoning}
& \multicolumn{2}{c}{Chinese}
& \multicolumn{1}{c}{Multilingual} \\
\cmidrule(lr){4-6}
\cmidrule(lr){7-8}
\cmidrule(lr){9-9}
\cmidrule(lr){10-11}
\cmidrule(lr){12-12}
(Param. Ratio) & & PPL & MMLU
& MMLU-Pro
& SuperGPQA
& MATH
& GSM8K
& BBH
& C-Eval
& CMMLU
& MMMLU \\
\midrule
None (0\%)
& 1.202 & 5.55 & \bf{68.25} & 44.38 & 25.22 & 46.02 & \bf{74.79} & \bf{65.71} & 70.78 & 73.01 & \bf{58.32} \\
5$\times$ (10\%)
& 1.200 & \bf{5.54} & 68.15 & 44.49 & 24.25 & 45.64 & 72.86 & 65.39 & 70.93 & \bf{73.44} & 57.49 \\
10$\times$ (25\%)
& \bf{1.197} & 5.55 & 67.71 & \bf{44.66} & \bf{25.64} & \bf{46.56} & 73.65 & 65.54 & 70.71 & 73.31 & 56.22 \\
30$\times$ (50\%)
& 1.201 & 5.59 & 67.75 & 42.61 & 24.18 & 44.62 & 74.45 & 65.55 & \bf{72.49} & 73.28 & 56.66 \\
\bottomrule
\end{tabular}
}
\end{table*}
\paragraph{Allocation under Fixed Parameter Budget.}
Following prior work, we scale the N-gram embedding slots while reducing the number of experts to maintain a fixed total parameter budget. 
The results are reported in Table~\ref{tab:ngram-vocab-size-samebudget}. 
The loss varies non-monotonically with vocabulary size and is lowest at $10\times$ ($25\%$), consistent with the allocation sweet spots reported in prior work~\citep{liu2026scalingembeddings,cheng2026engram}.
The same optimum is not evident in other evaluations. The out-of-domain uncheatable PPL changes little across budgets, and downstream benchmarks show no clear improvement over the MoE-only baseline.
These results suggest that \textit{N-gram embeddings and MoE experts play distinct roles in scaling capacity.
We therefore study N-gram vocabulary scaling while holding the MoE parameter budget fixed.}

\paragraph{Scaling with Additional Parameters.}

\begin{table*}[h]
\centering
\caption{Effect of N-gram vocabulary scaling. Vocabulary scales are measured relative to the base tokenizer vocabulary size (250K). Larger vocabularies increase the total number of parameters.}
\label{tab:ngram-vocab-size}
\resizebox{\textwidth}{!}{
\begin{tabular}{l c ccc cc c cc c}
\toprule
& 
& \multicolumn{3}{c}{Knowledge}
& \multicolumn{2}{c}{STEM}
& \multicolumn{1}{c}{Reasoning}
& \multicolumn{2}{c}{Chinese}
& \multicolumn{1}{c}{Multilingual} \\
\cmidrule(lr){3-5}
\cmidrule(lr){6-7}
\cmidrule(lr){8-8}
\cmidrule(lr){9-10}
\cmidrule(lr){11-11}

Vocab. Scale & Loss & MMLU
& MMLU-Pro
& SuperGPQA
& MATH
& GSM8K
& BBH
& C-Eval
& CMMLU
& MMMLU \\
\midrule
None
& 1.585 & 62.78 & 33.43 & 20.97 & 32.52 & 59.21 & 53.40 & 66.91 & 68.10 & 54.06 \\
20$\times$
& 1.553 & 64.14 & 34.46 & 21.83 & \bf{37.38} & \bf{65.09} & 57.13 & 71.75 & 72.29 & 55.94 \\
50$\times$
& 1.541 & 64.71 & 35.80 & 21.49 & 37.32 & 64.00 & \bf{57.56} & 72.12 & 72.48 & 56.64 \\
100$\times$
& 1.534 & 64.70 & \bf{35.87} & \bf{21.93} & 36.98 & 63.08 & 56.03 & 73.75 & 72.73 & \bf{56.65} \\
200$\times$
& \bf{1.526} & \bf{64.85} & 35.21 & 21.11 & 35.34 & 62.96 & 56.23 & \bf{74.94} & \bf{73.24} & 55.82 \\
\bottomrule
\end{tabular}
}
\end{table*}

Because embedding tables are sparsely accessed and deterministically addressed, they can be scaled with negligible additional per-token computation and stored in off-accelerator storage~\citep{google2025gemma3n,cheng2026engram}.
We therefore move beyond the fixed-size setting and scale the N-gram vocabulary from $20V$ to $200V$, where $V$ denotes the base tokenizer vocabulary size of Qwen3.5~\citep{qwenteam2026qwen35}.
The results are shown in Table~\ref{tab:ngram-vocab-size}.
Loss decreases monotonically as the N-gram vocabulary grows, while downstream performance does not follow the same trend.
Vocabulary scaling yields broad gains over the baseline, but performance on some benchmarks saturates or fluctuates as the vocabulary grows.
Furthermore, performance on Chinese benchmarks (e.g., C-Eval and CMMLU) improves consistently with N-gram vocabulary size.

Beyond vocabulary scaling, we explored a range of strategies for improving parameter efficiency of N-gram embeddings, including but not limited to token normalization for vocabulary compression~\citep{cheng2026engram}, non-uniform allocation across N-gram orders, and frequency-based partitioning of embedding slots.
Despite these efforts, we observed no consistent performance gains in our training recipe.

\section{Optimization}

\subsection{Optimizer}
\label{sec:optimizer}

\paragraph{Motivation.}
\label{sec:optimizer-motivation}


The matrix-based optimizer Muon~\citep{jordan2024muon} computes the update direction for a matrix parameter by orthogonalizing the momentum using Newton--Schulz (NS) iterations, and has been shown effective at scale~\citep{liu2025moonlight,kimi2025k2}. We use Muon as the main optimizer and find that several design choices influence its practical efficiency and stability, which we describe below.

\paragraph{Orthogonalization.}
\label{sec:optimizer-ns}


The NS iteration is applied to a Nesterov-accelerated momentum~\citep{jordan2024muon} with $\mu = 0.95$, and the orthogonalized result is scaled by $\gamma(A,B) = 0.2\sqrt{\max(A,B)}$ for a parameter of shape $A\times B$, making the RMS of the update independent of the matrix shape~\citep{liu2025moonlight}.
For the NS iteration, we adopt the per-step coefficient schedule of the Polar Express method~\citep{amsel2025polarexpress}, which is minimax-optimal for a given step budget. 
We set the number of iteration steps to $8$, which provides more accurate orthogonalization than fewer steps and reduces both the magnitude and frequency of gradient-norm spikes in our stress test.
We set the numerical stability constant in the Frobenius normalization preceding the NS iteration to $10^{-14}$.

\paragraph{Which Parameters Use Muon.}
\label{sec:optimizer-assignment}


We apply Muon to the two-dimensional weights that genuinely act as linear maps: the attention q/k/v and output projections, the GDN~\citep{yang2024gated} input and output projections, the fc1/fc2 of both routed and shared experts, and the key/value projection in N-gram embedding layers.
The input embeddings and the output head stay on AdamW.
For the MoE router, we observe that Muon exacerbates early-training fluctuations and destabilizes the router. Although applying Muon to the router during the mid-to-late training stages does not cause instability, we find no significant performance gains. Therefore, we use AdamW for the router.
One possible explanation is that each output dimension of the router corresponds to the score of one expert, and the dimensions are largely independent, leaving no shared linear structure for orthogonalization to exploit.
The two low-rank projections of GR (\S\ref{sec:residual}) likewise perform better with AdamW.
We attribute this to their very elongated shape.
Finally, the $n$-gram embedding table runs on Adam with weight decay disabled.

\paragraph{Splitting Fused Parameters.}
\label{sec:optimizer-split}
In Megatron-LM~\citep{shoeybi2019megatron}, the attention $qkv$ projection, the SwiGLU~\citep{shazeer2020glu} fc1, and the GDN input projection are each stored as one fused matrix, but semantically they are concatenations of independent linear operators along the output dimension.
Orthogonalizing the fused matrix is then wrong in two ways: the iteration mixes singular directions across unrelated sub-blocks, and $\gamma(A,B)$ is computed from the concatenated shape instead of the true operator shape.
We therefore split the fused gradient before orthogonalization, run NS on each sub-matrix independently, and gather the results back into the original layout before applying the update.
The $qkv$ and GDN input projections are split at per-head granularity, which improves both loss and downstream benchmarks. 
The fc1 is split into its gate and up halves; the loss is essentially unchanged while benchmarks improve slightly.
Splitting also provides a natural granularity for excluding individual sub-matrices from Muon.
Specifically, the GDN decay and beta projections produce one scalar per head and are therefore vectors, on which orthogonalization is not meaningful. 
For the output gates (the attention output gate~\citep{qiu2025gated} and the GDN $z$ projection), our ablations found AdamW on par with or slightly better than Muon.

\paragraph{Implementation.}
\label{sec:optimizer-impl}
The NS iteration introduces two main implementation challenges for Muon:
First, the iteration requires a holistic update over each full parameter matrix ($\approx 4K\max(A,B)\min(A,B)^2$ FLOPs for $K$ iteration steps), which clashes with Megatron's sharding in two ways: under TP, no rank owns the full weight matrix; under DP, the cost is cubic in the shorter dimension, so Megatron's equal-element partition leaves severe stragglers. We developed \textbf{Canzona}, which decouples logical optimizer assignment from physical parameter layout~\citep{wang2026canzonaunifiedasynchronousloadbalanced}: an $\alpha$-balanced static partitioner reassigns whole parameters (no cut inside a tensor) to equalize estimated NS FLOPs across DP ranks, and an asynchronous Micro-Group pipeline reconstructs each Muon-owned matrix via fused All-to-All across TP ranks. 
Each owner then runs a step mathematically equivalent to single-device Muon, ZeRO-1's bucket geometry is preserved so Megatron's Reduce-Scatter/backward overlap is retained, and the abstraction extends to other matrix-based optimizers.
Second, after splitting, one layer contributes on the order of a hundred sub-matrices, and the optimizer step becomes a long sequence of very small kernels bounded by launch overhead rather than arithmetic. 
We capture the whole step in a CUDA graph to remove this overhead.

\subsection{Hyperparameter Scaling}
\label{sec:hpscaling}

\paragraph{Motivation.}
\label{sec:hpscaling-motivation}

Choosing near-optimal hyperparameters is critical for efficient and stable model training. In previous generations of Qwen models, we fitted scaling laws for key hyperparameters, such as the learning rate ($\eta$) and batch size ($B$) \citep{qwen25, yang2025qwen3}. The optimal learning rate and batch size depend on the model architecture and the optimizer, so a recipe that was optimal for the previous-generation model may become suboptimal once both change.
Under our previous Qwen3.5 hyperparameter recipe~\citep{qwenteam2026qwen35}, we find that the new architecture and optimizer train noticeably more stably than before (\S\ref{sec:stability}), suggesting room for a more aggressive hyperparameter setting to further improve training performance.

Therefore, we develop an updated hyperparameter scaling law.
The architecture and optimizer changes shift the predicted near-optimal hyperparameters toward substantially larger batch sizes and learning rates, with a slower decay in the learning rate as model size increases.

A scaling law fitted at small scales is only useful if it extrapolates reliably.
We validated the learning-rate and batch-size predictions separately, evaluating each in a regime designed to make its effect more pronounced: the batch size prediction on a small model trained for many tokens (where an excessively large batch size may lead to suboptimal performance), and the learning rate prediction on a large model trained for a limited token budget (where training instability is the primary risk).
The predicted hyperparameters lie in the near-optimal basin and yield clear improvements over the previous Qwen3.5 recipe in both pretraining loss and benchmark performance.

\begin{figure}[t]
  \centering
  \vskip -0.2in
  \begin{subfigure}[t]{0.49\linewidth}
    \centering
    \includegraphics[pagebox=cropbox,width=\linewidth]{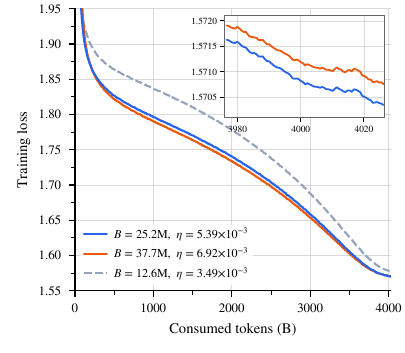}
    \vskip -0.05in
    \caption{Constant batch size.}
    \label{fig:hp-bsz-compare}
  \end{subfigure}
  \hfill
  \begin{subfigure}[t]{0.49\linewidth}
    \centering
    \includegraphics[pagebox=cropbox,width=\linewidth]{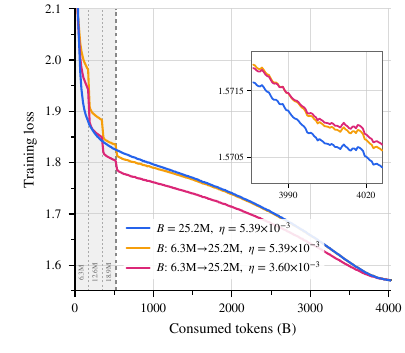}
    \vskip -0.05in
    \caption{Ramped batch size.}
    \label{fig:hp-bsz-warmup}
  \end{subfigure}
  \vskip -0.05in
  \caption{\textbf{Batch size on a 4T-token budget.} Training loss against consumed tokens for a 20-layer 10.8B-A0.89B MoE; each inset resolves the last 50B tokens. 
  \textbf{(a)} Moving from the previous recipe ($B\,{=}\,12.6$M) to the predicted optimum ($B\,{=}\,25.2$M) is worth $7.2\times10^{-3}$, while a further step to $B\,{=}\,37.7$M leads to a minor degradation. The loss rises steeply below the prediction and remains nearly flat above it.
  \textbf{(b)} Reaching $B\,{=}\,25.2$M through a ramp instead performs no better than using this batch size from the start and takes $18.8\%$ more optimizer steps.}
\end{figure}

\paragraph{Batch Size at a Large Token Budget.}
\label{sec:hpscaling-bsz}
The batch-size prediction is evaluated on a 20-layer 10.8B-A0.89B MoE model trained over $4$T tokens.
We compare the previous recipe ($B\,{=}\,12.6$M), the optimum predicted by the new scaling law ($B\,{=}\,25.2$M), and a setting $1.5\times$ larger ($B\,{=}\,37.7$M), with each configuration using the learning rate prescribed by the scaling law for its respective batch size.
All runs consume the same token budget, ensuring the comparison is based on equal compute.

Averaged over the final 20B tokens, the losses are $1.5702$ at $B\,{=}\,25.2$M , $1.5707$ at $B\,{=}\,37.7$M, and $1.5774$ under the previous recipe (Fig.~\ref{fig:hp-bsz-compare}).
The new scaling-law fit therefore improves the loss by $7.2\times10^{-3}$ over the previous recipe, while a further $1.5\times$ increase in batch size incurs a $4.3\times10^{-4}$ penalty, which is not significant.
The loss increases sharply below the predicted batch size and plateaus above it, indicating that the prediction is close to optimal and large enough to realize performance gains without being excessive.

\paragraph{Batch-Size Warmup Is No Longer Necessary.}
\label{sec:hpscaling-warmup}

Large runs commonly ramp the batch size over early training rather than starting at its final value~\citep{brown2020gpt3,liu2024deepseek}. 
The rationale is that the critical batch size is small early in training and grows over time~\citep{mccandlish2018largebatch,zhang2024criticalbatch}, making a large batch inefficient at the start. 
Furthermore, smaller batches paired with a scaled learning rate~\citep{goyal2017minibatch} are generally easier to stabilize during the initial steps.

However, our previous hyperparameter sweep suggests a different behavior when using the Muon optimizer. 
We observe that decreasing the batch size below the predicted optimum incurs a more significant performance penalty than increasing it, and that Muon preserves data efficiency at larger batch sizes where AdamW's performance degrades~\citep{jordan2024muon,essentialai2025muon}. 
For sparse MoE models, a larger batch also ensures that every expert receives a sufficient and diverse token signal per step, which plausibly aids expert specialization~\citep{global_balance}. 
Together, these observations led us to hypothesize that batch-size warmup may be unnecessary.

We therefore re-tested the batch-size warmup. 
We increase the batch size from $6.3$M in increments of $6.3$M, reaching the target of $25.2$M at 524B tokens. 
We evaluated two variants: the first maintains the peak learning rate of the constant-batch optimum, making the ramp its only difference; the second lowers the peak learning rate to account for the smaller batch sizes in the early stages.

Neither variant improves performance (Fig.~\ref{fig:hp-bsz-warmup}). 
Both ramps converge within the run-to-run variance but remain slightly worse than the constant-batch baseline, underperforming by $2.5\times10^{-4}$ and $3.5\times10^{-4}$, respectively. 
Additionally, the warmup requires $18.8\%$ more optimizer steps for the same token budget, resulting in greater wall-clock time overhead. 
Training stability is also unaffected: no run in this sweep exhibits a step where the loss exceeds its local median by more than $0.1$, and the p99.9 pre-clip gradient norm ranges from $0.088$ to $0.190$, well below the clipping threshold of $0.5$.

The loss trajectory of the warmup runs reveals the underlying mechanism. 
During the warmup phase, the smaller batch size introduces higher gradient noise under the same learning rate, resulting in a higher loss compared to the constant-batch baseline. 
Shortly after the batch size reaches its target, the warmup variant may exhibit a transient loss advantage due to the greater number of optimization steps accumulated during the early phase. 
However, as the learning rate decays and the model approaches convergence, this step-count advantage is neutralized. 
Ultimately, the constant-batch baseline surpasses the warmup runs, yielding a better final loss. 
Consequently, we do not employ batch-size warmup in our production runs.

\begin{figure}[h]
  \centering
  \begin{subfigure}[t]{0.49\linewidth}
    \centering
    \includegraphics[pagebox=cropbox,width=\linewidth]{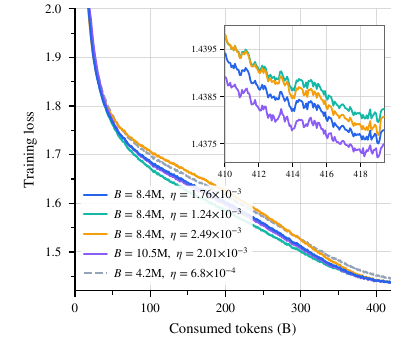}
    \vskip -0.05in
    \caption{Training loss.}
    \label{fig:hp-lr-loss}
  \end{subfigure}
  \hfill
  \begin{subfigure}[t]{0.49\linewidth}
    \centering
    \includegraphics[pagebox=cropbox,width=\linewidth]{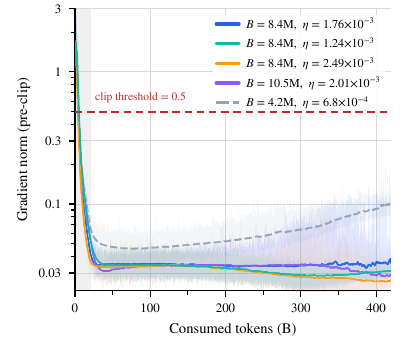}
    \vskip -0.05in
    \caption{Gradient norm.}
    \label{fig:hp-lr-gnorm}
  \end{subfigure}
  \vskip -0.05in
  \caption{\textbf{Learning rate at a larger model scale.} Five runs of a
    48-layer MoE on a 419B-token budget: the predicted optimum ($B\,{=}\,8.4$M,
    $\eta\,{=}\,1.76\times10^{-3}$), $\eta$ divided and multiplied by $\sqrt{2}$,
    a $25\%$ larger batch with its matched $\eta$, and the previous recipe
    (dashed). \textbf{(a)} The inset covers the last 10B tokens. The loss differences among the predicted optimum and the three nearby settings are near the noise level, so the predicted optimum sits at the bottom of a flat bowl. 
    \textbf{(b)} Pre-clip gradient norm on a log scale, with individual steps faint behind a moving average, the warmup shaded
    and the clip threshold dashed. After warmup, the pre-clip gradient norms of all runs near the predicted optimum remain below $50\%$ of the clipping threshold, including the run at $\sqrt{2}$ times the predicted learning rate.}
\end{figure}

\paragraph{Learning Rate at a Larger Model Scale.}
\label{sec:hpscaling-lr}
The learning-rate prediction is tested on a much larger $48$ layers 156B-A7B MoE, over a $419$B-token budget that is again the same for every run.
We evaluate the predicted optimum together with learning rates scaled by $1/\sqrt{2}$ and $\sqrt{2}$, a $25\%$ larger batch with its matched learning rate, and the previous hyperparameter recipe.


The previous recipe ends $7.8\times10^{-3}$ above the predicted optimum, while the four settings near the predicted optimum end within $7\times10^{-4}$ of each other, near the noise level (Fig.~\ref{fig:hp-lr-loss}).
The optimum therefore sits at the bottom of a bowl that is flat over at least a factor of $\sqrt{2}$ in either direction in learning rate and $+25\%$ in batch size; the larger batch is nominally the best of the four, by $3\times10^{-4}$.

\begin{table*}[t]
\vskip -0.2in
\centering
\small
\setlength{\tabcolsep}{3.5pt}
\caption{Downstream accuracy of the five learning-rate runs of
Fig.~\ref{fig:hp-lr-loss}, all at the same $419$B-token budget on the 48-layer 156B-A7B MoE. 
All values are percentages and higher is better; the benchmarks and the evaluation pipeline are those of \S\ref{sec:gdn-ablation}. 
Bold denotes the best result in each column.}
\vskip -0.05in
\label{tab:hp-lr-bench}
\resizebox{\textwidth}{!}{%
\begin{tabular}{lcccccccccc}
\toprule
& & & \multicolumn{3}{c}{Knowledge} & \multicolumn{2}{c}{STEM} & Reasoning & Multilingual & \\
\cmidrule(lr){4-6} \cmidrule(lr){7-8} \cmidrule(lr){9-9} \cmidrule(lr){10-10}
Setting
& \multicolumn{1}{c}{$B$}
& \multicolumn{1}{c}{$\eta$}
& \multicolumn{1}{c}{MMLU}
& \multicolumn{1}{c}{MMLU-Pro}
& \multicolumn{1}{c}{SuperGPQA}
& \multicolumn{1}{c}{MATH}
& \multicolumn{1}{c}{GSM8K}
& \multicolumn{1}{c}{BBH}
& \multicolumn{1}{c}{MMMLU}
& \multicolumn{1}{c}{Avg.} \\
\midrule
New fit, predicted optimum
& $8.4$M & $1.76\times10^{-3}$
& \textbf{73.84} & 48.35 & \textbf{29.31} & \textbf{49.98} & \textbf{80.89} & 73.25 & 68.23 & \textbf{60.55} \\
New fit, $\eta\div\sqrt{2}$
& $8.4$M & $1.24\times10^{-3}$
& 73.59 & 47.00 & 28.10 & 48.92 & 77.48 & 72.00 & 66.88 & 59.14 \\
New fit, $\eta\times\sqrt{2}$
& $8.4$M & $2.49\times10^{-3}$
& \textbf{73.84} & 46.92 & 28.04 & 49.58 & 80.06 & \textbf{73.82} & \textbf{68.46} & 60.10 \\
New fit, $B\times1.25$
& $10.5$M & $2.01\times10^{-3}$
& 73.73 & \textbf{48.51} & 28.52 & 49.32 & 80.06 & 72.58 & 67.72 & 60.06 \\
Qwen3.5 recipe
& $4.2$M & $6.8\times10^{-4}$
& 71.23 & 45.35 & 25.67 & 45.54 & 74.32 & 69.54 & 63.19 & 56.41 \\
\bottomrule
\end{tabular}
}
\end{table*}

Downstream benchmark results demonstrate the robust convergence achieved by our new scaling-law fit (Tab.~\ref{tab:hp-lr-bench}). 
The predicted optimum yields the highest average accuracy, securing the best or tied-best scores across the majority of evaluated tasks, while the previous recipe falls noticeably behind. 
Crucially, increasing the batch size or learning rate beyond the predicted optimum results in only a marginal, statistically insignificant drop in benchmark performance. 
This indicates a highly stable optimization landscape where the model's generalization does not sharply degrade with slight hyperparameter deviations. 
We treat these specific rankings as observational; given the single evaluation per run and the narrow margins among the top settings, these minor variations likely fall within standard evaluation noise.

Beyond final performance, maintaining training stability is paramount at large model scales.
The training dynamics remain exceptionally stable across all configurations. 
Gradient clipping never engages after the warmup phase in any of the five runs. 
At the predicted optimum, the maximum pre-clip gradient norm reaches only $28\%$ of the clipping threshold, compared to $51\%$ under the previous recipe, which suffers from noisier gradients due to its smaller batch size (Fig.~\ref{fig:hp-lr-gnorm}). 
Furthermore, the loss curves are remarkably smooth without any loss spike; no run exhibits a single step where the loss exceeds its local median by more than $0.1$. 
Even further increasing the predicted learning rate at this scale remains entirely stable. 
This confirms that our new scaling law does not dangerously over-extrapolate, providing an efficient, safe and robust hyperparameter recipe for large-scale training.

\begin{figure}[h]
  \centering
  \begin{subfigure}[t]{0.49\linewidth}
    \centering
    \includegraphics[pagebox=cropbox,width=\linewidth]{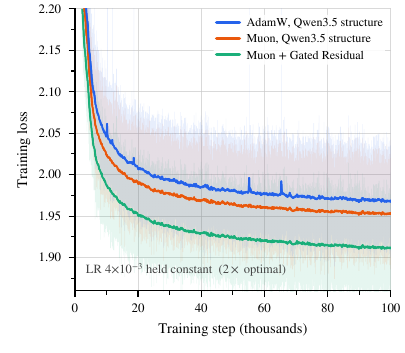}
    \vskip -0.05in
    \caption{$2\times$ optimal learning rate.}
    \label{fig:stress-2x}
  \end{subfigure}
  \hfill
  \begin{subfigure}[t]{0.49\linewidth}
    \centering
    \includegraphics[pagebox=cropbox,width=\linewidth]{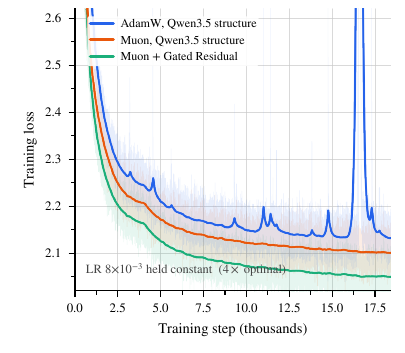}
    \vskip -0.05in
    \caption{$4\times$ optimal learning rate.}
    \label{fig:stress-4x}
  \end{subfigure}
  \vskip -0.05in
  \caption{\textbf{Training loss under stress.} The 28-layer 25B-A3B MoE at a constant learning rate: the Qwen3.5 structure under AdamW, the same structure under Muon, and Muon with GR. Bold lines are a moving average over the faint per-step trace. GR enables more stable training.}
  \label{fig:stress-loss}
\end{figure}

\subsection{Stability Stress Test}
\label{sec:stability}

\paragraph{Motivation.}
\label{sec:stability-motivation}
When a model scales to trillions of parameters and is trained on tens of trillions of tokens, it enters a regime where stability challenges emerge that are entirely absent in smaller-scale experiments~\citep{chowdhery2023palm,zhang2022opt,dehghani2023vit22b,qwenteam2026qwen35}. 
For example, long training runs can spend substantially more optimizer steps at peak learning rates, increasing the opportunity for instabilities to emerge. 
Such instabilities often manifest as loss spikes or divergence that require checkpoint restarts.
To iterate efficiently at moderate scale while still surfacing the instabilities that would appear at production scale, we design a set of stress tests that amplify the relevant stress within a smaller budget.
This verification becomes essential when the architecture and optimizer change together, as they do in Qwen3.8-Flash-Next: the gated residual, the GDN hybrid, and Muon all alter how updates and activations are scaled, and confirming that these changes remain stable under prolonged training is a prerequisite for reliable scaling.

\paragraph{Stress Test Design.}
\label{sec:stability-method}
Following the observation that large-scale instabilities can be reproduced in small models by raising the learning rate~\citep{wortsman2023smallscale}, we hold the learning rate constant at a multiple of its optimal value, bypassing the standard decay schedule to simulate the prolonged peak learning rate of a production run. We apply this to a 28-layer MoE at $2\times$ and $4\times$ its optimal learning rate. The evaluation criterion is that the new recipe must be at least as stable as the previous Qwen3.5 structure with AdamW~\citep{qwenteam2026qwen35}, which has already been scaled successfully. 
All runs share the same batch size and a gradient-norm clipping threshold of $0.5$~\citep{pascanu2013clipping}. 
We measure three quantities: loss spikes (steps exceeding a $201$-step rolling median by more than $0.1$), the
$p_{99.9}$ of the pre-clip gradient norm and the number of threshold crossings, and the per-block maximum activation.

\paragraph{Stress Test Results.}
\label{sec:stability-stress}
The stress tests reveal a clear stability margin for the new recipe. 
At $2\times$ the optimal learning rate, the AdamW baseline begins to spike ($4.3$ per $10$k steps), while both Muon configurations remain highly stable ($0.2$ per $10$k steps). 
At $4\times$ the optimal learning rate, the differences become categorical (Fig.~\ref{fig:stress-loss}). 
The AdamW run spikes on $183$ per $10$k steps and crosses the clipping threshold on $213$ of $19{,}932$ steps, engaging the clipper continuously. 
In contrast, both Muon runs never cross the clipping threshold, and the configuration with the gated residual records zero loss spikes. 
Under equal stress, the new architecture and optimizer combination is substantially more stable.

\begin{figure}[t]
  \centering
  \begin{subfigure}[t]{0.49\linewidth}
    \centering
    \includegraphics[pagebox=cropbox,width=\linewidth]{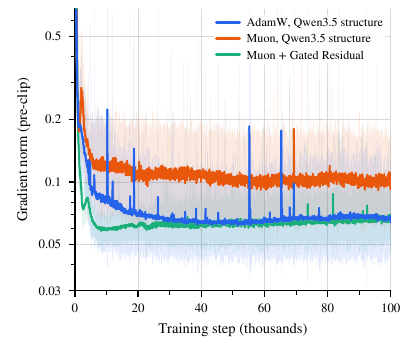}
    \vskip -0.15in
    \caption{Pre-clip gradient norm.}
    \label{fig:stress-gnorm}
  \end{subfigure}
  \hfill
  \begin{subfigure}[t]{0.49\linewidth}
    \centering
    \includegraphics[pagebox=cropbox,width=\linewidth]{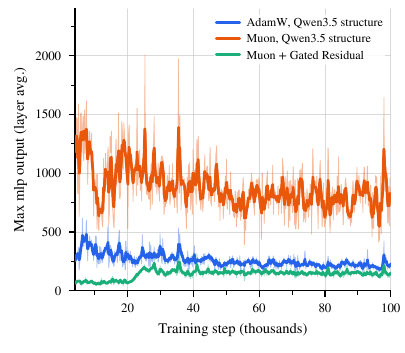}
    \vskip -0.15in
    \caption{Maximum MLP output.}
    \label{fig:stress-actmlp}
  \end{subfigure}
  \vskip -0.05in
  \caption{\textbf{Gradient norm and activations under stress.} 
    The same three runs as Fig.~\ref{fig:stress-loss} (a). 
    The activation in (b) is averaged over layers. Gradient Residual reduces both the frequency and magnitude of gradient-norm spikes, as well as the magnitude of activation outliers (Fig.~\ref{fig:stress-mechanism}).}
  \label{fig:stress-mechanism}
\end{figure}

\begin{figure}[h]
  \centering
  \begin{subfigure}[t]{0.33\linewidth}
    \centering
    \includegraphics[pagebox=cropbox,width=\linewidth]{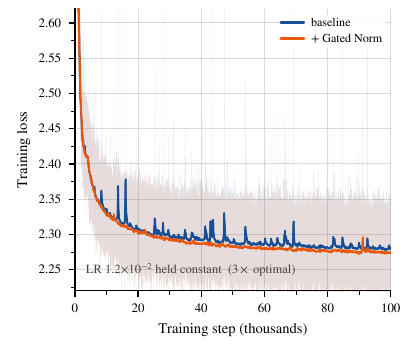}
    \vskip -0.05in
    \caption{Training loss.}
    \label{fig:gate-loss}
  \end{subfigure}
  \begin{subfigure}[t]{0.33\linewidth}
    \centering
    \includegraphics[pagebox=cropbox,width=\linewidth]{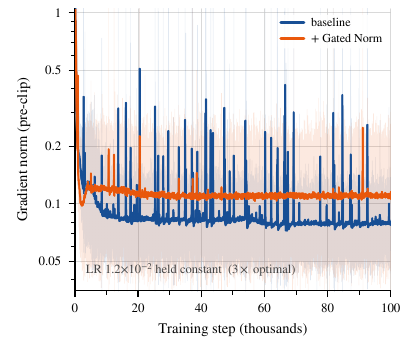}
    \vskip -0.05in
    \caption{Pre-clip gradient norm.}
    \label{fig:gate-gnorm}
  \end{subfigure}
  \begin{subfigure}[t]{0.33\linewidth}
    \centering
    \includegraphics[pagebox=cropbox,width=\linewidth]{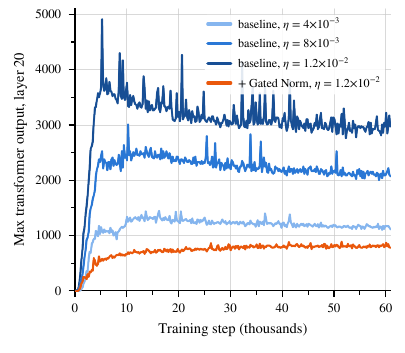}
    \vskip -0.05in
    \caption{Outliers against learning rate.}
    \label{fig:gate-ladder}
  \end{subfigure}
  \vskip -0.05in
  \caption{\textbf{Isolating the effect of the gate.} GatedNorm off and on, with the AdamW optimizer, structure and data order held fixed.
    \textbf{(a)} and \textbf{(b)} are the pair at $3\times$ the optimal learning rate; bold lines are a moving average over the faint per-step trace. \textbf{(c)} adds the ungated
    baseline at $1\times$ and $2\times$ the optimal rate.}
  \label{fig:gate}
\end{figure}

\paragraph{Mechanism: The Role of the Gate.}
\label{sec:stability-mechanism}
Analyzing the gradient norms and activations provides insight into this stability margin. 
At $2\times$ the optimal learning rate, Muon runs exhibit a higher median gradient norm and larger maximum activations than AdamW, yet they produce far fewer loss spikes. 
Adding GR reduces both the frequency and magnitude of gradient-norm spikes, as well as the magnitude of activation outliers (Fig.~\ref{fig:stress-mechanism}).

To isolate the effect of the gate, we evaluate a single-variable pair on the 28-layer model at $3\times$ its optimal learning rate, keeping the AdmaW optimizer and structure fixed while toggling GatedNorm (Fig.~\ref{fig:gate}). 
Enabling the gate reduces the spike rate from $32.0$ to $3.2$ per $10$k steps and cuts threshold crossings from $256$ to $20$. 
A learning-rate ladder on the ungated baseline shows that activation outliers grow almost proportionally with the learning rate, while the spike rate grows much faster (Fig.~\ref{fig:gate-ladder} (c)). 
With the gate enabled at the highest learning rate, the outlier level drops below the baseline at the lowest learning rate. 
This suggests that training at high learning rates requires a rescaling mechanism: without an explicit gate, the network achieves this by growing activation outliers, leaving it fragile; the multiplicative gate supplies the necessary rescaling directly, keeping the training  stable~\citep{qiu2025gated,qiu2026unified}.

\begin{figure}[h]
  \centering
  \begin{subfigure}[t]{0.9\linewidth}
    \centering
    \includegraphics[pagebox=cropbox,width=\linewidth]{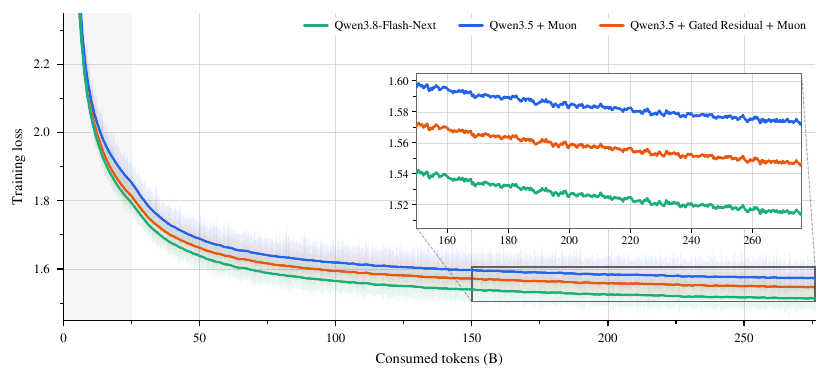}
    \vskip  -0.1in
    \caption{Training loss.}
    \label{fig:early-loss}
  \end{subfigure}
  \begin{subfigure}[t]{0.45\linewidth}
    \centering
    \includegraphics[pagebox=cropbox,width=\linewidth]{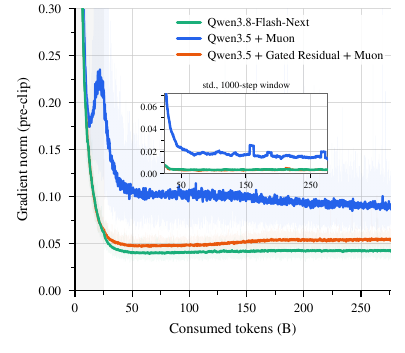}
    \vskip  -0.1in
    \caption{Pre-clip gradient norm and sliding window std.}
    \label{fig:early-gnorm}
  \end{subfigure}
  \vspace{4pt}
  \begin{subfigure}[t]{0.45\linewidth}
    \centering
    \includegraphics[pagebox=cropbox,width=\linewidth]{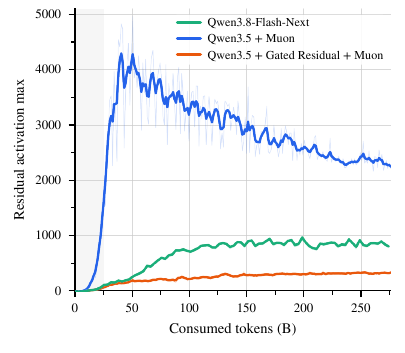}
    \vskip -0.05in
    \caption{Residual maximum over training.}
    \label{fig:early-actmax}
  \end{subfigure}
  \vskip -0.05in
  \caption{\textbf{The early phase of Qwen3.8-Flash-Next, at the shipped learning rate.} 
    The first 276B tokens of three runs that share data order, learning-rate schedule, and optimizer: Qwen3.5 with Muon, the same plus the gated residual, and Qwen3.8-Flash-Next.
    The shaded band corresponds to the learning-rate warmup. The inset in (a) resolves the last $126$B tokens of the window; the inset in (b) is the standard deviation of the gradient norm
    inside a rolling $1000$-step window. 
    }
  \label{fig:early}
\end{figure}

\paragraph{Verification at the Production Run.}
\label{sec:stability-production}
While the stress test amplifies instabilities by leaving the shipped
configuration, we also verify that the stability benefits persist under the
actual production training configuration. We compare the first $276$B tokens
of three runs sharing the same data order, learning-rate schedule, and
optimizer: the Qwen3.5 structure with Muon, the same structure plus GR, and
the full Qwen3.8-Flash-Next recipe with the further refined GR and the n-gram
embedding layer.

On loss (Fig.~\ref{fig:early}a), adding GR lowers the loss at $276$B tokens
by $0.026$, and the full Flash-Next recipe lowers it by a further
$0.032$, for a total gain of $0.058$ over the Muon baseline. This loss
improvement translates into significant benchmark gains
(\S\ref{sec:pt_eval}), enabling Qwen3.8-Flash-Next to reach pre-training
results comparable to Qwen3.7-Plus at roughly a ninth of the training
cost.

On gradient norm (Fig.~\ref{fig:early}b), Muon on its own has roughly twice the median norm and $4.2\times$ the $p_{99.9}$ of either gated run ($0.097$/$0.298$ vs.\ $0.053$/$0.071$ and $0.043$/$0.066$), and is the only
run to cross the clipping threshold. 
The gated runs are also steadier, with $4.3$--$4.7\times$ lower standard deviation inside a $1000$-step window, reproducing the stress-test finding at $8\times$ the model scale and at the production learning rate. 
Fusing the residual read and the final normalization before the LM head into one gated read operation further
reduces the gradient norm, likely the main contributor to the gap between Flash-Next and the Muon + GR configuration.
On activations (Fig.~\ref{fig:early}c), adding GR markedly reduces the
residual maximum throughout the network, consistent at every probed depth.
This allows stable training without explicit activation control such as qk-clip~\citep{kimi2025k2} and SwiGLU-clip~\citep{agarwal2025gpt}.

\section{Evaluation}


\label{sec:pt_eval}
We evaluate Qwen3.8-Flash-Next-Base against a set of strong base models across a broad range of capabilities, including general knowledge, reasoning, mathematics, scientific knowledge, coding, and multilingual understanding. The evaluation covers 14 benchmarks:

\begin{itemize}

\item \textbf{General Tasks}: MMLU~\citep{mmlu} (5-shot), MMLU-Pro~\citep{wang2024mmlu} (5-shot, CoT), MMLU-Redux~\citep{mmluredux} (5-shot), BBH~\citep{suzgun2022bbh} (3-shot, CoT), SuperGPQA~\citep{supergpqa} (5-shot, CoT).

\item \textbf{Math \& STEM Tasks}: GPQA~\citep{reinhardt2023gpqa} (5-shot, CoT), GSM8K~\citep{cobbe2021training} (4-shot, CoT), MATH~\citep{hendrycks2021measuring} (4-shot, CoT).

\item \textbf{Coding Tasks}: EvalPlus~\citep{evalplus} (0-shot; the average over HumanEval~\citep{humaneval}, MBPP~\citep{austin2021mbpp}, HumanEval+ and MBPP+), MultiPL-E~\citep{cassano2023multipl} (0-shot; Python, C++, Java, PHP, TypeScript, C\#, Bash, JavaScript), SWEBench-Pretrain, a pre-training variant of SWE-bench~\citep{swebench}\footnote{We draw inspiration from \citet{xia2024agentless} to construct SWEBench-Pretrain. In detail, we feed base models the problem descriptions, relevant code files and ask models to produce diff patches. The sequence similarities between golden and model-generated patches are calculated as the reported scores.}.

\item \textbf{Multilingual Tasks}: MGSM~\citep{shi2022language} (8-shot, CoT), MMMLU~\citep{openai2024mmmlu} (5-shot), INCLUDE~\citep{include} (5-shot).

\end{itemize}

\begin{table}[h]
\centering
\caption{Comparison among the base models of Qwen3.8-Flash-Next, Qwen3.8-27B and Qwen3.7-Plus. The highest and second-best scores are shown in \textbf{bold} and \underline{underlined}, respectively.}
\label{tab:qwen37-comparison}
\small
\begin{tabular}{@{}lccc@{}}
\toprule
 & \textbf{Qwen3.8-Flash-Next-Base} & \textbf{Qwen3.8-27B-Base} & \textbf{Qwen3.7-Plus-Base} \\
\midrule
\# Params & 125B & 27B & 397B \\
\# Activated Params & 6B & 27B & 17B \\
\# N-gram Embedding Params & 51B & -- & -- \\
\midrule
\multicolumn{4}{c}{\textit{General Tasks}} \\
\midrule
MMLU & \underline{90.36} & 87.51 & \textbf{90.43} \\
MMLU-Redux & \underline{90.68} & 87.26 & \textbf{91.47} \\
MMLU-Pro & \textbf{73.23} & 68.60 & \underline{70.90} \\
SuperGPQA & \textbf{51.36} & 44.86 & \underline{48.42} \\
BBH & \textbf{90.87} & 89.56 & \underline{89.41} \\
\midrule
\multicolumn{4}{c}{\textit{Math \& STEM Tasks}} \\
\midrule
GPQA & \underline{51.42} & 45.01 & \textbf{51.52} \\
GSM8K & \textbf{93.29} & \underline{93.18} & 92.95 \\
MATH & \underline{72.78} & 60.54 & \textbf{74.38} \\
\midrule
\multicolumn{4}{c}{\textit{Coding Tasks}} \\
\midrule
EvalPlus & \textbf{78.76} & 76.05 & \underline{78.06} \\
MultiPL-E & \underline{79.09} & 74.50 & \textbf{81.68} \\
SWEBench-Pretrain & \textbf{50.99} & 41.66 & \underline{49.24} \\
\midrule
\multicolumn{4}{c}{\textit{Multilingual Tasks}} \\
\midrule
MGSM & \textbf{89.33} & \underline{86.37} & 85.42 \\
MMMLU & \textbf{84.86} & 79.74 & \underline{84.53} \\
INCLUDE & \underline{78.40} & 74.37 & \textbf{78.90} \\
\bottomrule
\end{tabular}
\end{table}

Tab.~\ref{tab:qwen37-comparison} compares Qwen3.8-Flash-Next-Base with two strong baselines, Qwen3.8-27B-Base and Qwen3.7-Plus-Base. 
Qwen3.8-Flash-Next-Base consistently outperforms Qwen3.8-27B-Base across all 14 benchmarks, demonstrating gains across general knowledge, reasoning, mathematics, coding, and multilingual capabilities. 
More notably, it outperforms the much larger Qwen3.7-Plus-Base on 8 of 14 benchmarks while remaining competitive on the others. These results are achieved with only about 1/3 of the activated parameters and 1/3 of the training tokens, corresponding to roughly 1/9 of the training FLOPs. 

Beyond training efficiency, the architectural improvements of Qwen3.8-Flash-Next, together with its substantially smaller number of activated parameters, also lead to significantly lower inference cost. Overall, Qwen3.8-Flash-Next-Base delivers a substantially better performance-efficiency trade-off in both training and inference.



\section{Conclusion}

We have described the architecture of Qwen3.8-Flash-Next and the ablations that selected it. The resulting model retains the quality of the previous generation's 397B-A17B flagship while activating a third of the parameters, training on a third of the tokens, and consuming roughly a ninth of the FLOPs.

The design reflects a conviction that architecture, efficiency, and optimization form one coupled system. 
GR supplies a rescaling that markedly improves training stability, and that stability margin shifts the optimal learning rate and batch size upward, improving both throughput and convergence. 
Phase-by-phase cost accounting directed the sparse-attention design into the indexer and concentrated the gated residual's expressiveness on the read. 
Removing any axis from the loop would have admitted seemingly harmless shortcuts: sparse writes that degrade after post-training, positional encoding that appears dispensable during pre-training, or a batch-size warmup that costs extra optimizer steps for no gain.

Equally important is how these decisions were validated. Every claim was tested at a scale where the full evaluation budget remains tractable, and the settings were designed to surface the failure modes of production training: the stress test holds the learning rate at multiples of its optimal value so that instabilities emerge within a moderate budget, and scaling-law predictions are verified at the regime where each is most sensitive. Where pre-training metrics agreed but late-stage evaluation diverged, the joint protocol caught the discrepancy before it reached production. Looking forward, the tightest bottleneck is evaluation throughput: \textit{a cheaper mid-scale probe that reliably predicts post-training ordering would make the design space far more searchable.}

\section{Authors}

\textbf{Core Contributors:} Zihan Qiu, Zekun Wang, Xiao Li, Yanpeng Li, Yang Xu, Yixuan Wang, Huaqing Zhang, Rui Men, Bo Zheng\textsuperscript{\Letter}, Dayiheng Liu\textsuperscript{\Letter} 

\renewcommand{\thefootnote}{\Letter}
\footnotetext{Corresponding authors.}
\renewcommand{\thefootnote}{\arabic{footnote}}

\textbf{Contributors\footnote{Alphabetical order.}}: Bochao Mao, Chengruidong Zhang, Fan Zhou, Hao Luo, Haofeng Huang, Haoran Lian, Haoyan Huang, Hongqing Chen, Jianwei Zhang, Jing Xu, Junjie Wang, Langshi Chen, Liangyu Wang, Linlang Jiang, Man Yuan, Minmin Sun, Peng Jin, Siqi Zhang, Siyu Wang, Xingzhang Ren, Yakai Wang, Yi Zhang, Yiming Dong, Yizhong Cao, Yubo Ma, Yunfei Mao

\bibliography{custom}
\bibliographystyle{colm2024_conference}



\end{document}